\documentclass[conference]{IEEEtran}
\IEEEoverridecommandlockouts

\usepackage{cite}
\usepackage{amsmath,amssymb,amsfonts}
\usepackage{textcomp}
\usepackage{xcolor}

\usepackage{algorithm}
\usepackage{algpseudocodex}
\usepackage{nicefrac}       
\usepackage{microtype}      
\usepackage{xcolor}         
\usepackage{amsmath}
\usepackage{amssymb}
\usepackage{bm}
\usepackage{bbm}
\usepackage{todonotes}
\usepackage{multicol}
\usepackage{caption}
\usepackage{multirow}
\usepackage{tabularx}
\usepackage{booktabs}
\usepackage{tabularray}
\usepackage{makecell}
\usepackage{graphicx}           
\usepackage{duckuments}         
\usepackage{bold-extra}
\usepackage{wrapfig}
\usepackage{url}
\usepackage{hyperref}

\usepackage{caption}
\usepackage{subcaption}
\usepackage{svg}

\newcommand{\argmin}{\text{arg\,min}}

\algrenewcommand\algorithmicrequire{\textbf{Input:}}
\algrenewcommand\algorithmicensure{\textbf{Output:}}
\algnewcommand\algorithmicforeach{\textbf{for each}}
\algdef{S}[FOR]{ForEach}[1]{\algorithmicforeach\ #1\ \algorithmicdo}

\def\BibTeX{{\rm B\kern-.05em{\sc i\kern-.025em b}\kern-.08em
    T\kern-.1667em\lower.7ex\hbox{E}\kern-.125emX}}
\begin{document}

\title{Protecting Federated Learning from Extreme Model Poisoning Attacks via Multidimensional Time Series Anomaly Detection}

\thispagestyle{plain}
\pagestyle{plain}

\author{Edoardo Gabrielli, Dimitri Belli, Zoe Matrullo, Vittorio Miori, Gabriele Tolomei
\thanks{
    Edoardo Gabrielli (corresponding author) is with the Department of Computer, Control and Management Engineering, Sapienza University of Rome, Italy. E-mail: edoardo.gabrielli@uniroma1.it.
    
    Dimitri Belli and Vittorio Miori are with the National Research Council in Pisa, Italy.
    
    Zoe Matrullo is with the Department of Statistics, Ludwig-Maximilians-University of Munich, Germany.
    
    Gabriele Tolomei is with the Department of Computer Science, Sapienza University of Rome, Italy.

    This work has been submitted to the IEEE for possible publication. Copyright may be transferred without notice, after which this version may no longer be accessible.
}
}


\maketitle

\begin{abstract}
Current defense mechanisms against model poisoning attacks in federated learning (FL) systems have proven effective up to a certain threshold of malicious clients. 
\\
In this work, we introduce FLANDERS, a novel pre-aggregation filter for FL resilient to large-scale model poisoning attacks, i.e., when malicious clients far exceed legitimate participants.
FLANDERS treats the sequence of local models sent by clients in each FL round as a matrix-valued time series. Then, it identifies malicious client updates as outliers in this time series by comparing actual observations with estimates generated by a matrix autoregressive forecasting model maintained by the server.
Experiments conducted in several non-iid FL setups show that FLANDERS significantly improves robustness across a wide spectrum of attacks when paired with standard and robust existing aggregation methods.
\end{abstract}


\newcommand{\Prob}{\mathbb{P}}
\newcommand{\R}{\mathbb{R}}
\newcommand{\Z}{\mathbb{Z}}
\newcommand{\E}{\mathbb{E}}
\newcommand{\insta}{\bm{x}}
\newcommand{\X}{X}
\newcommand{\dataset}{\mathcal{D}}
\newcommand{\train}{\dataset_{\text{train}}}
\newcommand{\test}{\dataset_{\text{test}}}
\newcommand{\features}{\mathcal{X}}
\newcommand{\labels}{\mathcal{Y}}
\newcommand{\hypspace}{\mathcal{H}}
\newcommand{\params}{\bm{\theta}}
\newcommand{\Params}{\bm{\Theta}}
\newcommand{\w}{\bm{\omega}}
\newcommand{\loss}{\ell}
\newcommand{\Loss}{\mathcal{L}}
\newcommand{\ind}{\mathbbm{1}}

\newcommand{\dimitri}[1]{\todo[inline,color=orange!60]{{\bf Dimitri:} #1}}
\newcommand{\edo}[1]{\todo[inline,color=teal!60]{{\bf Edoardo:} #1}}
\newcommand{\gabri}[1]{\todo[inline,color=red!60]{{\bf Gabri:} #1}}

\section{Introduction}
\label{sec:introduction}

Recently, {\em federated learning} (FL) has emerged as the leading paradigm for training distributed, large-scale, and privacy-preserving machine learning (ML) systems~\cite{mcmahan2017googleai,mcmahan2017aistats}. 
The core idea of FL is to allow multiple edge clients to collaboratively train a shared, global model without disclosing their local private training data.
A typical FL round involves the following steps: {\em(i)} the server randomly picks some clients and sends them the current, global model; {\em(ii)} each selected client locally trains its model with its own private data; then, it sends the resulting local model to the server;\footnote{Whenever we refer to global/local model, we mean global/local model {\em parameters}.} {\em(iii)} the server updates the global model by computing an \emph{aggregation function}, usually the average (FedAvg), on the local models received from clients.
This process goes on until the global model converges. 
\\
Although its advantages over standard ML, FL also raises security concerns~\cite{costa2022covert}. 
Here, we focus on \emph{untargeted model poisoning} attacks~\cite{bhagoji2019pmlr}, where an adversary attempts to tweak the global model weights 
by directly perturbing the local model's parameters of some infected clients before these are sent to the central server for aggregation.
In doing so, the adversary aims to jeopardize the global model \textit{indiscriminately} at inference time.
Such model poisoning attacks severely impact standard FedAvg; therefore, more robust aggregation functions must be designed to secure FL systems.
\\
Unfortunately, existing defense mechanisms either rely on simple heuristics (e.g., Trimmed Mean and FedMedian by~\cite{yin2018icml}) or need strong and unrealistic assumptions to work effectively (e.g., foreknowledge or estimation of the number of malicious clients in the FL system, as for Krum/Multi-Krum~\cite{blanchard2017nips} and Bulyan~\cite{mhamdi2018pmlr}, which, however, cannot exceed a fixed threshold).
Furthermore, outlier detection methods using K-means clustering~\cite{shen2016acm} or spectral analysis like DnC~\cite{shejwalkar2021ndss} do not directly consider the temporal evolution of local model updates received.
Finally, strategies like FLTrust~\cite{cao2020fltrust} require the server to collect its own dataset and act as a proper client, thereby altering the standard FL protocol.
\\
%
This work introduces a novel pre-aggregation \textit{filter} robust to untargeted model poisoning attacks. Notably, this filter $(i)$ operates without requiring prior knowledge or constraints on the number of malicious clients and $(ii)$ inherently integrates temporal dependencies. 
The FL server can employ this filter as a preprocessing step before applying \textit{any} aggregation function, be it standard like FedAvg or robust like Krum or Bulyan.
Specifically, we formulate the problem of identifying corrupted updates as a multidimensional (i.e., matrix-valued) time series anomaly detection task. 
The key idea is that legitimate local updates, resulting from well-calibrated iterative procedures like stochastic gradient descent (SGD) with an appropriate learning rate, show \textit{higher predictability} compared to malicious updates. This hypothesis stems from the fact that the sequence of gradients (thus, model parameters) observed during legitimate training exhibit regular patterns, as validated in Section~\ref{subsec:intuition}. 
\\
Inspired by the matrix autoregressive (MAR) framework for multidimensional time series forecasting~\cite{chen2021je}, we propose the FLANDERS ({\em \textbf{F}ederated \textbf{L}earning meets \textbf{AN}omaly \textbf{DE}tection for a \textbf{R}obust and \textbf{S}ecure}) filter.
The main advantages of FLANDERS over existing strategies like FLDetector~\cite{zhao2020multivariate} are its resilience to large-scale attacks, where $50\%$ or more FL participants are hostile, and the capability of working under realistic non-iid scenarios.
We attribute such a capability to two key factors: $(i)$ FLANDERS works without knowing a priori the ratio of corrupted clients, and $(ii)$ it embodies temporal dependencies between intra- and inter-client updates, quickly recognizing local model drifts caused by evil players. Below, we summarize our main contributions:

\begin{itemize}
\item[{\em(i)}]
We provide empirical evidence that the sequence of models sent by legitimate clients is more predictable than those of malicious participants performing untargeted model poisoning attacks.
\\
\item[{\em(ii)}] 
We introduce FLANDERS, the first pre-aggregation filter for FL robust to untargeted model poisoning based on multidimensional time series anomaly detection.
\\
\item[{\em(iii)}] 
We integrate FLANDERS into Flower,\footnote{\scriptsize{\url{https://flower.dev/}}} a popular FL simulation framework for reproducibility.
\\
\item[{\em(iv)}] 
We show that FLANDERS improves the robustness of the existing aggregation methods under multiple settings: different datasets, client's data distribution (non-iid), models, and attack scenarios.
\\
\item[{\em(v)}] 
We publicly release all the implementation code of FLANDERS along with our experiments.\footnote{\scriptsize{\url{https://anonymous.4open.science/r/flanders_exp-7EEB}}}
\end{itemize}

The remainder of the paper is structured as follows. 
Section~\ref{sec:background} covers background and preliminaries. 
In Section~\ref{sec:related}, we discuss related work.
Section~\ref{sec:problem} and Section~\ref{sec:method} describe the problem formulation and the method proposed. 
Section~\ref{sec:experiments} gathers experimental results. 
Finally, we conclude in Section~\ref{sec:conclusion}.

\section{Background and Preliminaries}
\label{sec:background}
\newcommand\addtag{\refstepcounter{equation}\tag{\theequation}}
\subsection{Federated Learning}
\label{subsec:fl}
We consider a typical supervised learning task under a standard FL setting, which consists of a central server $S$ and a set of distributed clients $\mathcal{C}$, such that $|\mathcal{C}|=K$.
Each client $c\in \mathcal{C}$ has its own private training set $\dataset_c$, namely the set of its $n_c$ local labeled examples, i.e., $\dataset_c = \{\bm{x}_{c,i}, y_{c,i}\}_{i=1}^{n_c}$.
\\
The goal of FL is to train a global predictive model whose architecture and parameters $\params^*\in \R^d$ are shared across all clients by solving 
$\params^* = \text{argmin}_{\params} \Loss(\params) = \text{argmin}_{\params} \sum_{c=1}^K p_c \Loss_c(\params;\dataset_c),$
where $\Loss_c$ is the local objective function for client $c$. Usually, this is defined as the empirical risk calculated over the training set $\dataset_c$ sampled from the client's local data distribution:
$\Loss_c(\params;\dataset_c) = \frac{1}{n_c}\sum_{i=1}^{n_c} \loss(\params;(\insta_{c,i}, y_{c,i})),$
where $\loss$ is an instance-level loss, e.g., cross-entropy (classification) or squared error (regression). 
Each $p_c \geq 0$ specifies the relative contribution of each client. 
Since it must hold that $\sum_{c=1}^{K}p_c = 1$, two possible settings are: $p_c = 1/K$ or $p_c = n_c/n$, where $n = \sum_{c=1}^K n_c$. 

The generic federated round at each time $t$ is decomposed into the following steps and iteratively performed until convergence, i.e., for each $t\in \{1,2,\ldots,T\}$:

 \begin{enumerate}
     \item[{\em(i)}] $S$ randomly selects a subset of clients ${\mathcal{C}}^{(t)}\subseteq \mathcal{C}$, so that $1 \leq |{\mathcal{C}}^{(t)}| \leq K$, and sends them the current, global model $\params^{(t)}$. At $t=1$, $\params^{(1)}$ is randomly initalized.
     \item[{\em(ii)}] Each selected client $c\in {\mathcal{C}}^{(t)}$ trains its local model $\params_c^{(t)}$ on its own private data $\dataset_c$ by optimizing the following objective, starting from $\params^{(t)}$:
     \[
     \params_c^{(t)} = \text{argmin}_{\params^{(t)}}\Loss_c(\params^{(t)}; \dataset_c). \addtag
     \]
     The value of $\params_c^{(t)}$ is computed via gradient-based methods (e.g., stochastic gradient descent) and sent to $S$.
     \item[{\em(iii)}] $S$ computes $\params^{(t+1)} = \phi(\{\params_c^{(t)}~|~c\in \mathcal{C}^{(t)}\})$ as the updated global model, where $\phi: \R^{d^m} \mapsto \R^d $ is an \emph{aggregation function}; for example, $\phi = \frac{1}{m}\sum_{c\in \mathcal{C}^{(t)}} \params_c^{(t)}$, i.e., FedAvg or alike~\cite{lu2020spml}.
 \end{enumerate}



\subsection{The Attack Model: Federated Aggregation under Model Poisoning}
\label{subsec:byzantine}
The most straightforward aggregation function $\phi$ the server can implement is FedAvg, which computes the global model as the average of the local model weights received from clients. 
FedAvg is effective when all the FL participants behave honestly~\cite{dean2012nips,konecny2016nipsws,mcmahan2017aistats}.
Instead, this work assumes that an attacker controls a fraction $b=\lceil r*K \rceil, r\in [0,1]$ of the $K$ clients, i.e., $0\leq b\leq K$, known as malicious. This is in contrast to previous works, where $r < 0.5$.
Below, we describe our attack model.

\noindent{{\bf {\em Attacker's Goal.}}}
Inspired by many studies on poisoning attacks against ML~\cite{rubinstein2009imc,biggio2012icml,biggio2013icb,xiao2015icml,li2016nips,yang2017ndss,jagielski2018sp}, we consider the attacker's goal is to jeopardize the jointly learned global model \textit{indiscriminately} at inference time with \textit{any} test example. 
Such attacks are known as {\em untargeted}~\cite{fang2020usenix}, as opposed to {\em targeted} poisoning attacks, where instead, the goal is to induce prediction errors only for some specific test inputs (e.g., via so-called \textit{backdoor triggers} as shown by~\cite{bagdasaryan2020pmlr}).

\noindent{{\bf {\em Attacker's Capability.}}}
Like Sybil attacks to distributed systems~\cite{douceur2002iptps}, the attacker can inject $b$ fake clients into the FL system or compromise $b$ honest clients.
The attacker can arbitrarily manipulate the local models sent by malicious clients to the server $S$.
More formally, let $x \in \mathcal{C}$ be one of the $b$ corrupted clients selected by the server on the generic $t$-th FL round; it first computes its legitimate local model $\params_x^{(t)}$ without modifying its private data $\dataset_x$; then it finds $\check{\params}_x^{(t)}$ by applying a {\em post hoc} perturbation $\bm{\varepsilon} \in \R^d$ to $\params_x^{(t)}$. For example, $\check{\params}_x^{(t)} = \params_x^{(t)} + \bm{\varepsilon}$, where $\bm{\varepsilon}\sim \mathcal{N}(\bm{\mu}, \bm{\Sigma})$ is a Gaussian noise vector. 
More advanced attack strategies have been designed, as discussed in Section~\ref{subsec:attacks}.

\noindent{{\bf {\em Attacker's Knowledge.}}}
We assume the attacker knows its controlled clients' code, local training datasets, and local models. 
Moreover, 
we consider the worst-case scenario, where the attacker is \textit{omniscient}, i.e., it has full knowledge of the parameters sent by honest parties. This allows the attacker to choose the best parameters to inject into the FL protocol, e.g., by crafting malicious local models that are close to legitimate ones, thus increasing the probability of being chosen by the server for the aggregation.

\noindent{{\bf {\em Attacker's Behavior.}}}
In each FL round, the attacker, similar to the server, randomly selects $b$ clients to corrupt out of the $K$ available. Any of these $b$ malicious clients that happen to be among those selected by the server will poison their local models using one of the strategies outlined in Section~\ref{subsec:attacks}. 
Note that, unlike earlier studies, we do not assume that malicious clients are chosen in the initial round and remain constant throughout the training. We argue that this approach is more realistic, as legitimate clients can be compromised at any point in actual deployments. Therefore, an effective defense in FL should ideally exclude these clients as soon as they submit their first malicious model.

This category of {\em untargeted model poisoning} attacks has been extensively explored in previous works~\cite{blanchard2017nips,bhagoji2019pmlr,fang2020usenix}.
It has been shown that including updates even from a single malicious client can wildly disrupt the global model if the server runs standard FedAvg~\cite{blanchard2017nips,yin2018icml}. 
\section{Related Work}
\label{sec:related}

Below, we describe the most popular defenses against model poisoning attacks on FL systems, which will serve as the baselines for our comparison.
A more comprehensive discussion is in~\cite{hu2021arxiv,barroso2023if}.
%

\noindent{{\bf {\em FedMedian}}~\cite{yin2018icml}{\bf.}}
The central server sorts the $j$-th parameters received from all the $m$ local models and takes the median of those as the value of the $j$-th parameter of the global model. This process is applied for all the model parameters, i.e., $\forall j\in \{1,\ldots,d\}$.
\\
\noindent{{\bf {\em Trimmed Mean}}~\cite{xie2018corr}{\bf.}}
This rule computes a model as FedMedian does, and then it averages the $k$ nearest parameters to the median.
If $b$ clients are compromised at most, this aggregation rule achieves an order-optimal error rate when $b \leq \frac{m}{2} - 1$.
\\
\noindent{{\bf {\em Krum}}~\cite{blanchard2017nips}{\bf.}}
It selects one of the $m$ local models received from the clients, which is most similar to \emph{all} other models, as the global model. The rationale behind this approach is that even if the chosen local model is poisoned, its influence would be restricted because it resembles other local models, potentially from benign clients.
A variant called Multi-Krum mixes Krum with standard FedAvg. 
%
\\
\noindent{{\bf {\em Bulyan}}~\cite{mhamdi2018pmlr}{\bf.}}
Since a single component can largely impact the Euclidean distance between two high-dimensional vectors, Krum may lose effectiveness in complex, high-dimensional parameter spaces due to the influence of a few abnormal local model weights. Bulyan iteratively applies Krum to select $\alpha$ local models and aggregates them with a Trimmed Mean variant to mitigate this issue.
\\
\noindent{{\bf {\em DnC}}~\cite{shejwalkar2021ndss}{\bf.}}
This robust aggregation 
uses spectral analysis to detect and filter outliers as proposed by~\cite{diakonikolas2017icml}. 
Similarly, DnC computes the principal component of the set of local updates sent by clients. Then, it projects each local update onto this principal component. Finally, it removes a constant fraction of the submitted model updates with the largest projections.
\\
\noindent{{\bf {\em FLDetector}}~\cite{zhang2022fldetector}{\bf.}}
This method is the closest to our approach. It filters out malicious clients by measuring their consistency across $N$ rounds using an approximation of the integrated Hessian to compute the \textit{suspicious scores} for all clients. However, unlike our method, FLDetector struggles with large numbers of malicious clients, and cannot work with highly heterogeneous data.

\section{Problem Formulation}
\label{sec:problem}

\subsection{Time Series of Local Models}
\label{subsec:ts-models}
At the end of each FL round $t$, the central server $S$ collects the updated local models $\{\params_c^{(t)}\}_{c\in \mathcal{C}^{(t)}}$ sent by the subset of selected clients.\footnote{A similar reasoning would apply if clients sent their local displacement vectors $\bm{u}_c^{(t)} = \params_c^{(t)} - \params^{(t)}$ or local gradients $\nabla{\Loss}_c^{(t)}$ rather than local model parameters $\params_c^{(t)}$.}
Without loss of generality, we assume that the number of clients picked at each round is constant and fixed, i.e., $|\mathcal{C}^{(t)}| = m,~\forall t\in \{1,2,\ldots, T\}$.
Hence, the server arranges the local models received at round $t$ into a $d\times m$ matrix $\Params_{t} = [\params_1^{(t)}, \ldots, \params_c^{(t)}, \ldots, \params_m^{(t)}]$, whose $c$-th column corresponds to the $d$-dimensional vector of updated parameters $\params_c^{(t)}$ sent by client $c$.
However, the subset of selected clients may be different at each FL round, i.e., $\mathcal{C}^{(t)} \neq \mathcal{C}^{(t')}$ for $t\neq t'$, although we have assumed their size ($m$) is the same. 
More generally, to track the local models sent by \textit{all} clients chosen across $1\leq w \leq T$ rounds, we can extend each $\Params_{t}$ into a $d\times h$ matrix, where $h = |\bigcup_{t=1}^w \mathcal{C}^{(t)}|$, such that $m \leq h \leq K$, with $K$ being the total number of clients. 
Notice that $h=m$ when $w=1$, whereas $h=K$ if the server selects all $K$ clients at least once over the $w$ rounds considered.
At every round $t$, $\Params_{t}$ contains $m$ columns corresponding to the local models sent by the clients \textit{actually} selected for that round. In contrast, the remaining $h-m$ ``fictitious'' columns refer to the unselected clients.
We fill these columns with the current global model at FL round $t$, i.e., $\params^{(t)}$, as if this were sent by the $h-m$ unselected clients. 
Note that this strategy is neutral and will not impact the aggregated global model computed by the server for the next round $(t+1)$, as this is calculated only from the updates received by the $m$ clients previously selected. 

\subsection{Predictability of Local Models Evolution: Legitimate vs. Malicious Clients}
\label{subsec:intuition}
\begin{figure}[ht!]
    \centering
    \includegraphics[width=\columnwidth]{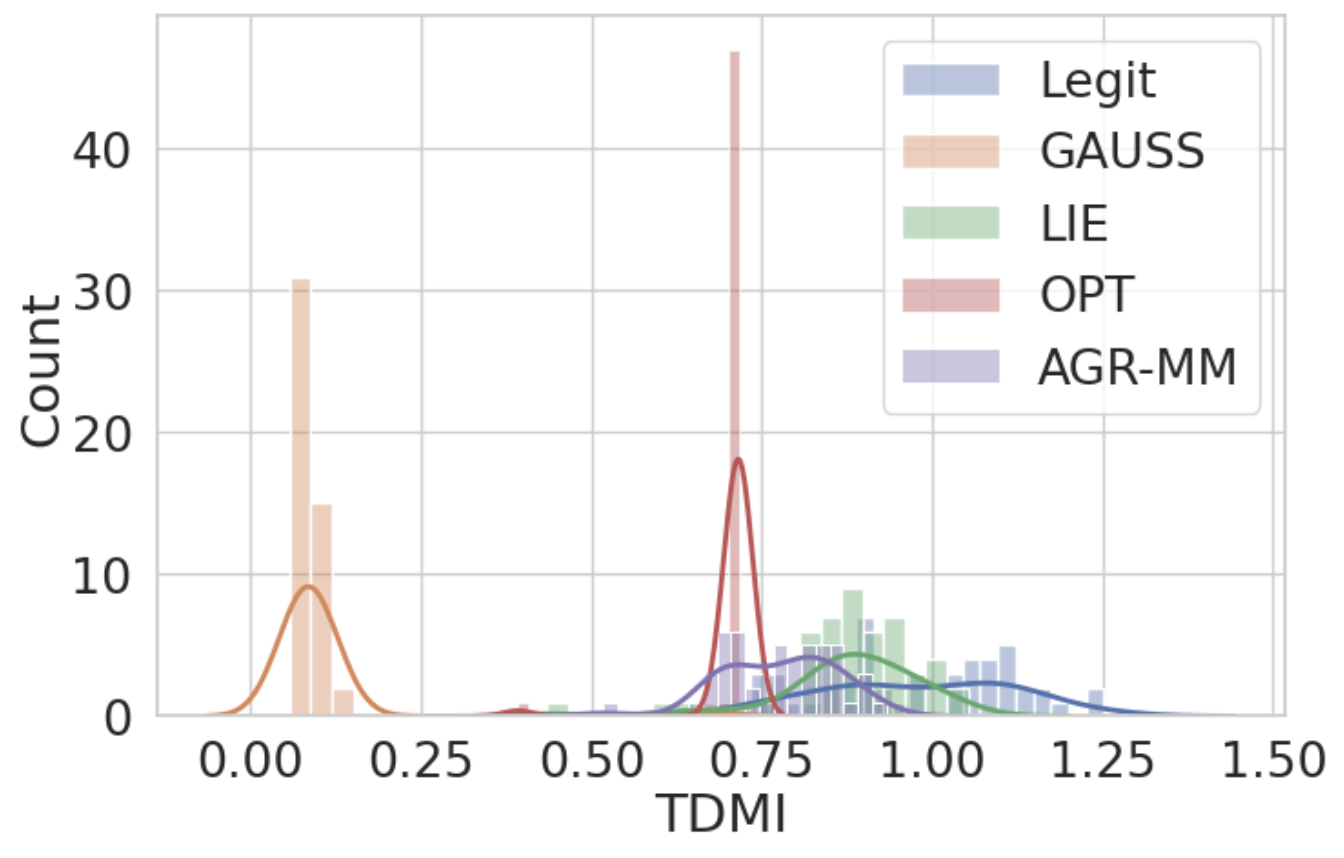}
    \caption{Empirical distributions of average TDMI computed between each pair of \textit{consecutive} local models sent by the legitimate client $i$ $(\params_i^{(t)}, \params_i^{(t+1)})$ and the malicious client $j$ $(\params_j^{(t)}, \params_j^{(t+1)})$, when this runs one of the four attacks considered in this work, namely GAUSS, LIE, OPT, and AGR-MM.}
    \label{fig:avg-tdmi-density}
\end{figure}

We claim that legitimate local updates 
show \textit{higher predictability} compared to malicious updates. 
This hypothesis stems from the fact that the sequence of gradients (hence, model parameters) observed during legitimate training should exhibit ``regular'' patterns until convergence. 
\\
To demonstrate this behavior, we consider two clients $i,j \in \mathcal{C}$. Client $i$ is assumed to be a legitimate participant, while client $j$ acts maliciously. 
Both are selected by the server over a sequence of consecutive $T$ FL rounds to train a global model on the \textit{MNIST} dataset.
Therefore, we examine the local models sent to the server at each round $t \in \{1,2,\ldots,T\}$ by clients $i$ and $j$. Specifically, these are $d$-dimensional vectors of real-valued parameters $\params_i^{(t)}, \params_j^{(t)} \in \R^d$, such that $\params_i^{(t)} = (\theta_{i,1}^{(t)}, \ldots, \theta_{i,d}^{(t)})$ and $\params_j^{(t)} = (\theta_{j,1}^{(t)}, \ldots, \theta_{j,d}^{(t)})$, respectively.
Next, we calculate the average time-delayed mutual information (TDMI) for each pair of observed local models $(\params_i^{(t)}, \params_i^{(t')})$ and $(\params_j^{(t)}, \params_j^{(t')})$ across $T=50$ rounds, where $\delta > 0$, for both client $i$ and $j$. This analysis aims to discern the time-dependent nonlinear correlation and the level of predictability between observations. 
We expect the TDMI calculated between pairs of models sent by the legitimate client $i$ to be higher than that computed between pairs sent by the malicious client $j$. Thus, the temporal sequence of local models observed from client $i$ is more predictable than that of client $j$.
\\
Let $\params_c^{(t)} = (\theta_{c,1}^{(t)}, \ldots, \theta_{c,d}^{(t)})$ denote the local model sent by the generic FL client $c\in \mathcal{C}$ as an instance of a $d$-dimensional vector-valued process. 
We define $\text{TDMI}(\params_c^{(t)}, \params_c^{(t')})$ as follows: 
\begin{equation}
\footnotesize
\text{TDMI}(\params_c^{(t)}, \params_c^{(t')}) =  \int p(\params_c^{(t)}, \params_c^{(t')})\times \log\Bigg(\frac{p(\params_c^{(t)}, \params_c^{(t')})}{p(\params_c^{(t)})p( \params_c^{(t')})}\Bigg) d\params_c^{(t)} d\params_c^{(t')},
\label{eq:tdmi2}
\end{equation}

where $p(\cdot)$ is the probability density function and $t'=t+\delta$. 
If each realization $\theta_{c,k}^{(t)}$ in the vector $\params_c^{(t)}$ is \textit{independent} from each other, we can compute the average TDMI as follows~\cite{albers2012chaos}:
\begin{equation}
\small
\text{Avg}\Big[\text{TDMI}(\params_c^{(t)}, \params_c^{(t')})\Big] = \frac{1}{d}\Bigg[\sum_{k=1}^d \text{TDMI}(\theta_{c,k}^{(t)}, \theta_{c,k}^{(t')})\Bigg], 
\label{eq:avg-tdmi}
\end{equation}
where TDMI$(\theta_{c,k}^{(t)}, \theta_{c,k}^{(t')})$ calculates TDMI for the univariate case.
Indeed, the $d$ model parameters $\theta_{c,k}^{(\cdot)}$ should reasonably be independent. Under this assumption, the joint probability density function of the parameters factors into a product of individual probability density functions, forming a product measure on $d$-dimensional Euclidean space. Thus, according to Fubini's theorem~\cite{billingsley2017probability}, the integral of each parameter will be independent of the others because the parameters are independent.
\\
To mimic the behavior of a hypothetical optimal FLANDERS filter, and therefore avoid the propagation of poisoned local models sent by client $j$ across the $T$ rounds, we assume that, at each round $t$, the global model from which both clients start their local training process is polished from any malicious updates received at the previous round $t-1$.
Next, we calculate the average time-delayed mutual information (TDMI) for each pair of observed local models $(\params_i^{(t)}, \params_i^{(t')})$ and $(\params_j^{(t)}, \params_j^{(t')})$ across $T=50$ rounds, where $t' > t$, for both client $i$ and $j$.
Firstly, we consider the special case where $t'=t+1$ and compute the average TDMI between each pair of \textit{consecutive} local models sent by the legitimate client and the malicious client, when this runs one of the four attacks considered in this work, namely GAUSS, LIE, OPT, and AGR-MM presented in Section~\ref{subsec:attacks}.
In Figure~\ref{fig:avg-tdmi-density}, we plot the empirical distributions of the observed average TDMI for the legitimate and malicious clients. 
To validate our claim that legitimate models are more predictable than malicious ones, we compute the mean of the empirical distributions for the legitimate and malicious client, $\bar{\params_i}$ and $\bar{\params_j}^{atk}$, respectively, where $atk=\{$GAUSS, LIE, OPT, AGR-MM$\}$. Then, we run a one-tailed $t$-test against the null hypothesis $H_0: \bar{\params_i} = \bar{\params_j}^{atk}$, where the alternative hypothesis is $H_a: \bar{\params_i} > \bar{\params_j}^{atk}$. 
The results of these statistical tests are illustrated in Table~\ref{tab:t-test}, showing that, in all four cases, there is enough evidence to reject the null hypothesis at a confidence level $\alpha=0.01$.


\begin{table}[htb!]
\centering
\caption{One-tailed $t$-test against the null hypothesis $H_0: \bar{\boldsymbol{\theta}_i} = \bar{\boldsymbol{\theta}_j}^{atk}$. Each cell contains the $p$-value for the statistical test corresponding to a specific attack. In all four cases, there is enough evidence to reject the null hypothesis at a confidence level $\alpha=0.01$ ($p$-value $\ll 0.01$).}
\label{tab:t-test}
\begin{tabular}{cc}
    \toprule
    & \rule{0pt}{3pt} $H_0: \bar{\boldsymbol{\theta}_i} = \bar{\boldsymbol{\theta}_j}^{atk}$ \\
    \midrule
    \multicolumn{1}{c}{GAUSS} & $5.67*10^{-38}$ \\
    \midrule
    \multicolumn{1}{c}{LIE} & $8.33*10^{-5}$ \\
    \midrule
    \multicolumn{1}{c}{OPT} & $6.70*10^{-18}$ \\
    \midrule
    \multicolumn{1}{c}{AGR-MM} & $4.20*10^{-12}$ \\
    \bottomrule
\end{tabular}
\end{table}



\subsection{Poisoned Local Models as Matrix-Based Time Series Outliers}
\label{subsec:ts-outliers}
We, therefore, formulate our problem as a multidimensional time series anomaly detection task.
At a high level, we want to equip the central server with an anomaly scoring function that estimates the degree of each client picked at the current round being malicious, i.e., its {\em anomaly score}, based on the historical observations of model updates seen so far from the clients selected.
Such a score will be used to restrict the set of trustworthy candidate clients for the downstream aggregation method. 
Note that unselected clients will \textit{not} contribute to the aggregation; thus, there is no need to compute their anomaly score even if they were marked as suspicious in some previous rounds.
On the other hand, we must design a fallback strategy for ``cold start'' clients  -- i.e., FL participants who send their local updates for the first time or have not been selected in any of the previous rounds considered  -- whether honest or malicious.
\\
More formally, let $\hat{\Params}_t = f(\Params_{t-w:t-1};\hat{\bm{\Omega}})$ be the matrix of local model updates \textit{predicted} by the central server $S$ at the generic FL round $t$. 
The forecasting model $f$ depends on a set of parameters $\hat{\bm{\Omega}}$ estimated using the past $w$ model updates observed, i.e., $\Params_{t-w:t-1}$, where $1\leq w \leq t-1$. 
Also, the server sees the actual matrix $\Params_t$.
\\
The anomaly score $s_{c}^{(t)} \in \R$ of the generic client $c$ at round $t$ can thus be defined as follows:
\begin{equation}
\small
\label{eq:anomaly-score}
s_{c}^{(t)} =
\begin{cases}
\delta(\params^{(t)}_c, \hat{\params}^{(t)}_c),\text{ if }c\in \mathcal{C}^{(t)} \wedge \exists j\in\{1..w\}\text{ s.t. }c\in \mathcal{C}^{(t-j)} \\
\delta(\params^{(t)}, \params^{(t)}_c), \text{ if }c\in \mathcal{C}^{(t)} \wedge \nexists j\in\{1..w\}\text{ s.t. }c\in \mathcal{C}^{(t-j)}\\
\perp, \text{ if }c\notin \mathcal{C}^{(t)}.
\end{cases}
\end{equation}
The first condition refers to a client selected for round $t$, which also appeared at least once in the history. 
In this case, $\delta$ measures the distance between the observed vector of weights sent to the server ($\params^{(t)}_c$) and the predicted vector of weights ($\hat{\params}^{(t)}_c$) output by the forecasting model $f$.
The second condition, instead, occurs when a client is selected for the first time at round $t$ or does not appear in the previous $w$ historical matrices of observations used to generate predictions. Here, due to the cold start problem, we cannot rely on $f$ and, therefore, the most sensible strategy is to compute the distance between the current global model and the local update received by the new client. 
Finally, the anomaly score is undefined ($\perp$) for any client not selected for round $t$.
\\
The server will thus rank all the $m$ selected clients according to their anomaly scores (e.g., from the lowest to the highest).
Several strategies can be adopted to choose which model updates should be aggregated in preparation for the next round, i.e., to restrict from the initial set $\mathcal{C}^{(t)}$ to another (possibly smaller) set $\mathcal{C}_{*}^{(t)}\subseteq \mathcal{C}^{(t)}$ of trusted clients. 
For example, $S$ may consider only the model updates received from the top-$k$ clients ($1 \leq k \leq m$) with the smallest anomaly score, i.e., $\mathcal{C}_*^{(t)} = \{c\in \mathcal{C}^{(t)}~|~s_{c}^{(t)} \leq s_k^{(t)}\}$, where $s_k^{(t)}$ indicates the $k$-th smallest anomaly score at round $t$. 
Alternatively, the raw anomaly scores computed by the server can be converted into well-calibrated probability estimates. 
Here, the server sets a threshold $\rho \in [0,1]$ and aggregates the weights only of those clients whose anomaly score is below $\rho$, i.e., $\mathcal{C}_*^{(t)} = \{c\in \mathcal{C}^{(t)}~|~s_c^{(t)}\leq \rho\}$. 
In the former case, the number of considered clients ($k$) is bound apriori,\footnote{This may not be true if anomaly scores are not unique; in that case, we can simply enforce $|\mathcal{C}_*^{(t)}| = k$.} whereas the latter does not put any constraint on the size of final candidates $|\mathcal{C}_*^{(t)}|$.
Eventually, $S$ will compute the updated global model $\params^{(t+1)} = \phi(\{\params_c^{(t)}~|~c\in \mathcal{C}_*^{(t)}\})$, where $\phi$ is any aggregation function, e.g., FedAvg, Bulyan, or any other strategy.

Below, we describe how we use the matrix autoregressive (MAR) framework proposed by \cite{chen2021je} to implement our multidimensional time series forecasting model $f$, hence the anomaly score.

\section{Proposed Method: FLANDERS}
\label{sec:method}

\subsection{Matrix Autoregressive Model (MAR)}
\label{subsec:mar}
We assume the temporal evolution of the local models sent by FL clients at each round is captured by a matrix autoregressive model (MAR). 
In its most generic form, MAR($w$) is a $w$-order autoregressive model defined as follows:
\begin{equation}
\small
    \Params_t = {\bm A}_1 \Params_{t-1} {\bm B}_1 + \cdots + {\bm A}_w \Params_{t-w} {\bm B}_w + {\bm E}_t,
    \label{eq:mar-w}
\end{equation}
where $\Params_t$ is the $d\times h$ matrix of observations at time $t$, $\bm{\Omega} = \{{\bm A}_i, {\bm B}_i\}_{i=1}^w$ are $d\times d$ and $h\times h$ autoregressive coefficient matrices, and ${\bm E}_t$ is a $d\times h$ white noise matrix.
In this work, we consider the simplest MAR($1$)\footnote{Unless otherwise specified, whenever we refer to MAR, we assume MAR($1$).} Markovian forecasting model, i.e., ${\Params}_t = {\bm A} {\Params}_{t-1} {\bm B} + {\bm E}_t$, where the matrix of local updates at time $t$ depends only on the matrix observed at time step $t-1$, namely $w=1$.

Let $\widetilde{\Params}_t \approx \Params_t$ be the {\em predicted} matrix of observations at time $t$, according to a model $f$ parameterized by $\widetilde{\bm{\Omega}} = \{\widetilde{{\bm A}},\widetilde{{\bm B}}\}$, i.e., $\widetilde{\Params}_t = f(\Params_{t-1};\widetilde{\bm{\Omega}}) = \widetilde{{\bm A}} \Params_{t-1}\widetilde{{\bm B}}$.
If we have access to $l > 0$ historical observations, we can estimate the best coefficients ${\hat{\bm \Omega}} = \{\hat{{\bm A}}, \hat{{\bm B}}\}$ by solving the following objective: 
\begin{equation}
\small
    {\hat{\bm \Omega}} = \hat{{\bm A}}, \hat{{\bm B}} = \argmin_{\widetilde{{\bm A}}, \widetilde{{\bm B}}}
    \Big\{\sum_{j=0}^{l-1} ||{\Params}_{t-j} - \widetilde{{\bm A}}{\Params}_{t-j-1}\widetilde{{\bm B}}||^2_{\text F} \Big\},
    \label{eq:mar-opt}
\end{equation}
where $||\cdot||_{\text F}$ indicates the Frobenius norm. 
Note that $l$ impacts only the size of the training set used to estimate the optimal coefficients $\hat{\bm A}$ and $\hat{\bm B}$; it does not affect the order of the autoregressive model, which will remain MAR($1$) and \textit{not} MAR($l$).
The optimal coefficients $\hat{\bm A}$ and $\hat{\bm B}$ can thus be estimated via alternating least squares (ALS) optimization~\cite{koren2009ieeecomp}.
Further details are provided in Appendix~\ref{app:mar}.

\subsection{MAR-based Anomaly Score}
\label{subsec:mar-anomaly-score}
Our approach consists of two primary steps: $(i)$ \textit{MAR estimation} and $(ii)$ \textit{anomaly score computation}. 
\\
\noindent{{\bf {\em MAR Estimation.}}}
At the first FL round ($t=1$), the server sends the initial global model $\params^{(1)}$ to the set of $m$ selected clients $\mathcal{C}^{(1)}$ and collects from them the $d\times m$ matrix of updated models $\Params_1$. Hence, it computes the \textit{new} global model $\params^{(2)} = \phi(\{\params_{c}^{(1)}~|~c\in \mathcal{C}^{(1)}\})$, where $\phi = \text{Krum}$ or any other existing robust aggregation heuristic.
For any other FL round $t > 1$, the server can use the past $l > 0$ historical observations $\Params_{t-l:t-1}$ to estimate the best MAR coefficients $\hat{\bm{\Omega}} = \{\hat{\bm{A}},\hat{\bm{B}}\}$ according to~(\ref{eq:mar-opt}). 
In general, $1\leq l\leq t-1$; however, if we assume $l$ fixed at each round, the server will consider $\Params_{\max(1,t-l):t-1}$ past observations.
Again, independently of the value of $l$, MAR will learn to predict the current matrix of weights \textit{only} from the previously observed matrix.
Therefore, each matrix used for training $\{\Params_{t-j-1}\}_{j=0}^{l-1}$ has the same size $d\times m$, as it contains exclusively the $m$ updates received at the round $t-j-1$. 
Of course, employing a higher order MAR($w$) model with $w > 1$ would require extending each observed matrix to $d\times h$ ($m < h \leq K)$, as detailed in Section~\ref{subsec:ts-models}. This is to track the local updates sent by selected clients across multiple historical rounds.
\\
\noindent{{\bf {\em Anomaly Score Computation.}}}
At the generic FL round $t > 1$, we compute the anomaly score using the estimated MAR forecasting model as follows. 
Let $\Params_{t}$ be the matrix of observed weights. 
This matrix may contain one or more corrupted local models from malicious clients.
Then, we compute the $m$-dimensional anomaly score vector ${\bm s}^{(t)}$, where ${\bm s}^{(t)}[c] = s_c^{(t)}$, as in~(\ref{eq:anomaly-score}).
A critical choice concerns the function $\delta$ used to measure the distance between the observed vector of weights sent by each selected client and the vector of weights predicted by MAR.
In this work, we set $\delta (\bm{u}, \bm{v}) = ||\bm{u} - \bm{v}||_2^2$, where $||\cdot||_2^2$ is the squared $L^2$-norm. 
Other functions can be used (e.g., {\em cosine distance}), especially in high dimensional spaces, where $L^p$-norm with $p\in (0,1)$ has proven effective~\cite{aggarwal2001icdt}. 
Choosing the best $\delta$ is outside the scope of this work, and we leave it to future study.
\\
According to one of the filtering strategies discussed in Section~\ref{subsec:ts-outliers}, we retain only the $k$ clients with the smallest anomaly scores. The remaining $m-k$ clients are considered malicious; thus, they 
are discarded and do not contribute to the aggregation run by the server as if they were never selected.
\\
At the next round $t+1$, we may want to refresh our estimation of the MAR model, i.e., to update the coefficient matrices $\hat{{\bm A}}$ and $\hat{{\bm B}}$.
We do so by considering the latest observed $\Params_{t}$ and the other $l-1$ previous matrices of local updates, using the same sliding window of size $l$.
Since the observed matrix $\Params_{t}$ contains $m-k$ potentially malicious clients, we cannot use it as-is. Otherwise, \textit{if any of the spotted malicious clients is selected again at round $t+1$}, we may alter the estimation of $\hat{{\bm A}}$ and $\hat{{\bm B}}$ with possibly corrupted matrix columns.
To overcome this problem, we replace the original $\Params_{t}$ with $\Params'_{t}$ {\em before}, feeding it to train the new MAR model. 
Specifically, $\Params'_{t}$ is obtained from $\Params_{t}$ by substituting the $m-k$ anomalous columns either with the parameter vectors from the same clients observed at time $t-1$, which are supposed to be still legitimate \textit{or} the current global model.
The advantage of this solution is twofold. 
On the one hand, a client labeled as malicious at FL round $t$ would likely still be considered so at $t+1$ if it keeps perturbing its local weights, thus improving robustness.
On the other hand, our solution allows malicious clients to alternate legitimate behaviors without being banned, speeding up model convergence.
Notice that the two considerations above might not be valid if we updated the MAR model using the original, partially corrupted $\Params_{t}$. 
Indeed, in the first case, the 
distance between two successive poisoned models by the same client would reasonably be small. So, the client's anomaly score will likely drop to non-alarming values, thereby increasing the number of false negatives.
In the second case, the 
distance between a corrupted and a legitimate model would likely be large. 
Thus, a malicious client will maintain its anomaly score high even if it acts honestly, impacting the number of false positives.

\subsection{A Step-by-Step Example}
\label{subsec:example}
\begin{figure}[htb!]
    \centering
    \includegraphics[width=\columnwidth]{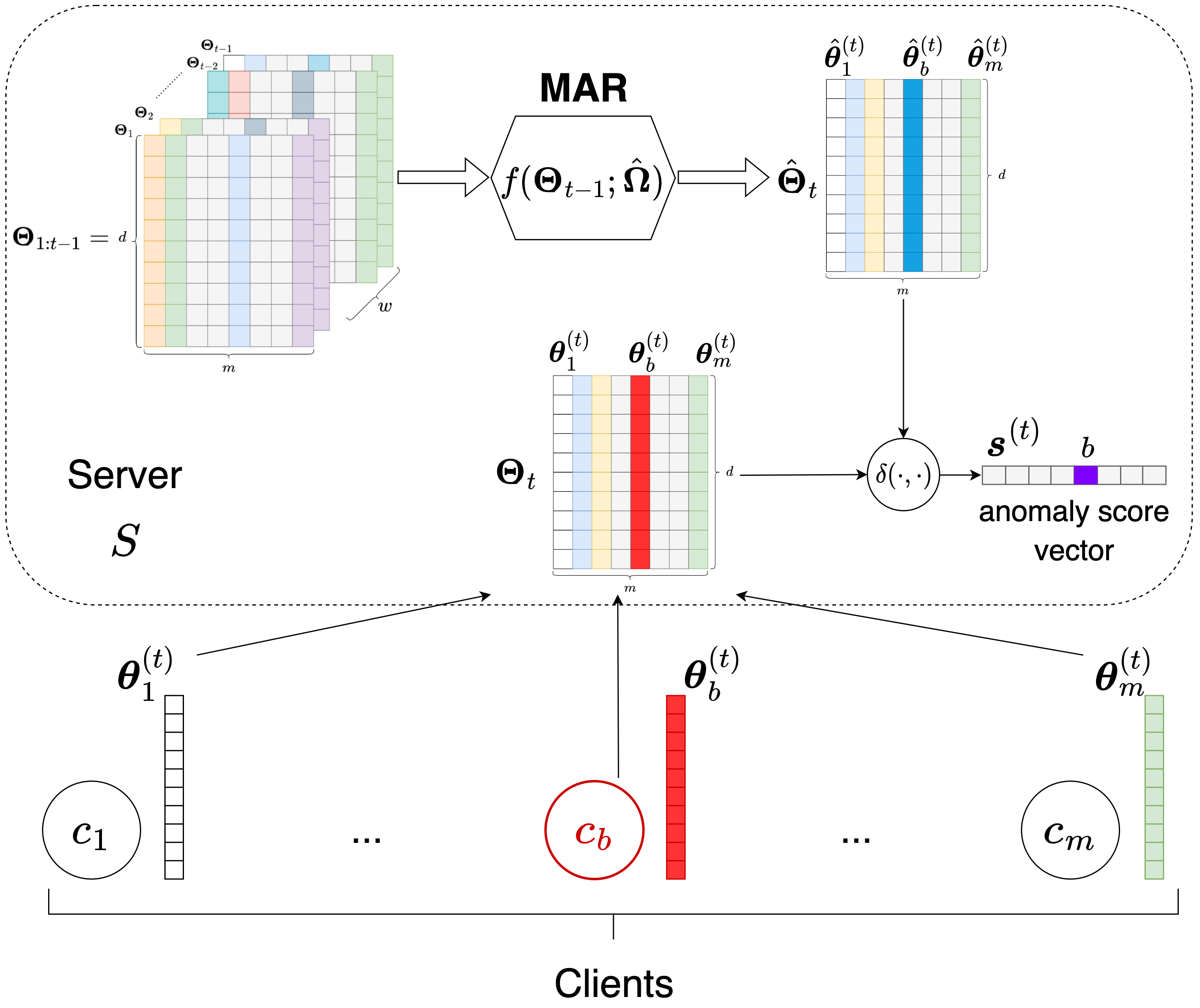}
    \caption{Overview of FLANDERS.}
    \label{fig:flanders}
\end{figure}

In Figure~\ref{fig:flanders}, we depict how FLANDERS computes the anomaly score vector $\bm{s}^{(t)}$ at the generic FL round $t$. Specifically, the server $S$ uses its current MAR($1$) forecasting model $f$ whose best parameters $\hat{\bm{\Omega}} = \{\hat{\bm A}, \hat{\bm B}\}$ are estimated from $l$ previous historical observations of local models received in the previous rounds $t-l,\ldots, t-1$. It applies $f$ to the previously observed matrix $\bm{\Params}_{t-1}$ to get the next predicted matrix of local models $\bm{\hat{\Params}}_{t}$, i.e., $\bm{\hat{\Params}}_{t} = f(\bm{\Params}_{t-1};\hat{\bm{\Omega}}) = \hat{\bm A}\bm{\Params}_{t-1} \hat{\bm B}$. Then, it compares this predicted matrix $\bm{\hat{\Params}}_{t}$ with the actual matrix of local updates received $\bm{\Params}_{t}$. 
The final anomaly score vector is calculated by measuring the distance $\delta$ between each column of those two matrices, according to Eq.~\ref{eq:anomaly-score}.

It is worth remarking that, in general, only some clients are selected at every round. In particular, if a client $c_{\text{new}}$ -- whether it is honest or malicious -- is selected by the server for the first time at round $t$, the MAR forecasting model $f$ will not be able to make any prediction for it due to a cold start problem (i.e., the predicted matrix $\bm{\hat{\Params}}_{t}$ will not contain a column corresponding to $c_{\text{new}}$). 
In such a case, we must adopt a fallback strategy. Without any historical information for a client, the most sensible thing to do is to compute the distance $\delta$ between the local model update it sent and the current global model.

%

To better clarify how FLANDERS works, consider the following practical example.
\\
Suppose an FL system consists of a centralized server and $10$ clients $c_1,\ldots, c_{10}$; furthermore, at each round, $4$ of those clients are randomly chosen for training. 
At the very first round ($t=1$), let $\mathcal{C}^{(1)} = \{c_2, c_3, c_7, c_9\}$ be the set of $4$ clients selected by the server. Hence, the server sends the current global model $\params^{(1)} \in \R^d$ to each of those clients and collects the $d\times 4$ matrix of updated local models $\Params_1 = [\params_2^{(1)}, \params_3^{(1)}, \params_7^{(1)}, \params_9^{(1)}]$. 
Therefore, it computes the \textit{new} global model $\params^{(2)} = \phi(\{\params_{c}^{(1)}~|~c\in \mathcal{C}^{(1)}\})$.
Notice that, at this stage, no anomaly score can be computed as FLANDERS cannot take advantage of any historical observations of local model updates. As such, if one (or more) selected clients in the very first round are malicious, plain FedAvg may not detect those. To overcome this problem, FLANDERS should be paired with one of the existing robust aggregation heuristics at $t=1$. For example, $\phi = \{\text{Trimmed Mean, Krum, Bulyan}\}$.

At the next round ($t=2$), FLANDERS can start using past observations (i.e., only $\Params_1$) to estimate the best MAR coefficients $\hat{\bm{\Omega}} = \{\hat{\bm{A}},\hat{\bm{B}}\}$ according to Eq.~(\ref{eq:app-mar-opt}) above. 
Suppose $\mathcal{C}^{(2)} = \{c_1, c_3, c_6, c_9\}$ is the set of $4$ clients selected by the server at the second round.
Let $\Params_2 = [\params_1^{(2)}, \params_3^{(2)}, \params_6^{(2)}, \params_9^{(2)}]$ be the local models sent by the clients to the server.
Moreover, $\hat{\Params}_2 = f(\Params_1;\hat{\bm{\Omega}}) = [\hat{\params}_2^{(2)}, \hat{\params}_3^{(2)}, \hat{\params}_7^{(2)}, \hat{\params}_9^{(2)}]$ are the local models predicted by MAR, using the previous set of observations. 

It is worth noting that $\mathcal{C}^{(1)} \cap \mathcal{C}^{(2)} = \{c_3, c_9\}$; the other two clients, $c_1$ and $c_6$, are considered ``cold-start'' since they are selected for the first time or, in any case, they do not appear in the historical observations used by MAR for predictions (i.e., in the previous $w=1$ matrices).
Thus, we calculate the following anomaly scores, according to Eq.~\ref{eq:mar-w}:
\begin{itemize}
\item $s_3^{(2)} = \delta(\params_3^{(2)}, \hat{\params}_3^{(2)})$ and $s_9^{(2)} = \delta(\params_9^{(2)}, \hat{\params}_9^{(2)})$, i.e., we measure the distance between the models actually sent by $c_3$ and $c_9$ and those predicted by MAR using the previous observations (\textit{first condition});
\item $s_1^{(2)} = \delta(\params^{(2)}, \params_1^{(2)})$ and $s_6^{(2)} = \delta(\params^{(2)}, \params_6^{(2)})$, i.e., we measure the distance between the models actually sent by $c_1$ and $c_6$ and the current global model at time $t=2$, as those clients were never picked before (\textit{second condition});
\item $s_2^{(2)} = s_4^{(2)} = s_5^{(2)} = s_7^{(2)} = s_8^{(2)} = s_{10}^{(2)} = \perp$, i.e., these clients were not selected at round $t=2$, and therefore they will not contribute to computing the new global model $\params^{(3)}$ anyway (\textit{third condition}).
\end{itemize}
Let us assume that $c_3$ is a malicious client controlled by an attacker. Moreover, suppose that this client started sending poisoned models since the very first round, i.e., $\params_3^{(1)}$ was already corrupted. 
In such a case, it is evident that if we had used plain FedAvg at $t=1$, this would have likely polluted the global model $\params^{(2)}$ and, therefore, FLANDERS might fail to recognize this as a malicious client due to a relatively low anomaly score $s_3^{(2)}$. As stated before, the consequences of this edge situation in which one or more malicious clients are picked at the beginning of the FL training can be mitigated by replacing FedAvg with one of the robust aggregation strategies available in the literature, such as Trimmed Mean, Krum, or Bulyan. 

However, suppose $c_3$ is correctly spotted as malicious at the end of round $t=2$ due to its high anomaly score $s_3^{(2)}$. Therefore, $c_3$ (actually, $\params_3^{(2)}$) will be discarded from the aggregation at the server's end. Hence, FedAvg can now safely be used; more generally, the server can restart running FedAvg from $t=2$ on, i.e., once FLANDERS can compute \textit{valid} anomaly scores.
The server can now compute the updated global model as $\params^{(3)} = \phi(\{\params_{c}^{(2)}~|~c\in \mathcal{C}_*^{(2)}\})$, where $\mathcal{C}_*^{(2)} \subset \mathcal{C}^{(2)}$ contains the clients with the $k$ smallest anomaly scores. 
For instance, if $k=2$ and $\mathcal{C}_*^{(2)} = \{c_1, c_9\}$, $\params^{(3)} = 1/2 * (\params_1^{(2)} + \params_9^{(2)})$.

At the next round ($t=3$), FLANDERS can use the previous two observations $\Params_1$ and $\Params_2$ to refine the estimation of the best MAR coefficients, again solving Eq.~(\ref{eq:app-mar-opt}).\footnote{In general, FLANDERS uses $l$ past observations $\Params_{\text{max}(1,t-l):t-1}$.} 
However, $\Params_2$ cannot be fed as-is to re-train MAR since one of its components -- i.e., the local model $\params_3^{(2)}$ sent by client $c_3$ -- has been flagged as malicious. 
Otherwise, the resulting updated MAR coefficients could be unreliable due to the propagation of local poisoned models.

To overcome this problem, FLANDERS replaces the local model marked as suspicious $\params_3^{(2)}$ with either one of the previously observed models from the same client that is supposedly legitimate \textit{or} the current global model. 
Since, in this example, we assume $c_3$ has been malicious from the first round, we use the latter approach. 
Specifically, we change $\Params_2$ with $\Params'_2 = [\params_1^{(2)}, \underline{\params_{}^{(2)}}, \params_6^{(2)}, \params_9^{(2)}]$, where $\params^{(2)}$ substitutes $\params_3^{(2)}$.
As discussed at the end of Section~\ref{subsec:mar-anomaly-score}, this fix allows FLANDERS to work even when a malicious client is picked for two or more rounds consecutively.

We can now compute $\hat{\Params}_3 = f(\Params'_2;\hat{\bm{\Omega}}) = [\hat{\params}_1^{(3)}, \hat{\params}_3^{(3)}, \hat{\params}_6^{(3)}, \hat{\params}_9^{(3)}]$ using the updated MAR, leveraging the previous set of amended observations. 
Let $\mathcal{C}^{(3)} = \{c_2, c_3, c_4, c_5\}$ denote the set of clients selected at round $3$ and $\Params_3 = [\params_2^{(3)}, \params_3^{(3)}, \params_4^{(3)}, \params_5^{(3)}]$ be the local models sent by the clients to the server. 
Thus, $\mathcal{C}^{(2)} \cap \mathcal{C}^{(3)} = \{c_3\}$. The remaining three clients -- $c_2$, $c_4$, and $c_5$ -- are treated as ``cold-start.'' Specifically, $c_4$ and $c_5$ are selected for the first time, while $c_2$, even though previously selected in the first round, was not chosen at the previous round ($2$). As $c_2$ was not part of the historical observations used by MAR($1$) to make predictions, we need to treat it as if it were a cold-start client.

Therefore, anomaly scores are updated as follows:
\begin{itemize}
\item $s_3^{(3)} = \delta(\params_3^{(3)}, \hat{\params}_3^{(3)})$, i.e., we measure the distance between the models actually sent by $c_3$ and those predicted by MAR using the previous observations (\textit{first condition});
\item $s_2^{(3)} = \delta(\params^{(3)}, \params_2^{(3)})$, $s_4^{(3)} = \delta(\params^{(3)}, \params_4^{(3)})$, and $s_5^{(3)} = \delta(\params^{(3)}, \params_5^{(3)})$ i.e., we measure the distance between the models actually sent by $c_2$, $c_4$, and $c_5$ and the current global model at time $t=3$, as those clients were never picked before or did not appear in the previous observations used to make predictions (\textit{second condition});
\item $s_1^{(3)} = s_6^{(3)} = s_7^{(3)} = s_8^{(3)} = s_9^{(3)} = s_{10}^{(3)} = \perp$, i.e., these clients were not selected at round $t=3$, and therefore they will not contribute to computing the new global model $\params^{(4)}$ anyway (\textit{third condition}).
\end{itemize}
Generally, the process above continues until the global model converges.

\subsection{FLANDERS' Pseudocode}
\label{subsec:pseudocode}
We present the pseudocode of a hypothetical FL server that integrates FLANDERS into the model aggregation stage. 
The main server-side loop is shown in Algorithm~\ref{alg:server}, whereas Algorithm~\ref{alg:flanders} details the computation of anomaly scores, which is at the heart of the FLANDERS filter.

\begin{algorithm}
    \caption{\texttt{\textsc{FLANDERS-Server}}}\label{alg:server}
    \begin{algorithmic}[1]
        \Require The aggregation function ($\phi$); the number of randomly selected clients at each FL round ($m$); the number of local models to keep as legitimate ($k$); the autoregressive order of MAR ($w$); the number of historical observations used to train MAR ($l$); the number of MAR training iterations ($N$); the number of total FL rounds ($T$).
        \Ensure The global model $\bm{\theta}^{(T)}$.%
        \Procedure{\texttt{FLANDERS-Server}}{$\phi$, $m$, $k$, $w$, $l$, $N$, $T$}
            \State $\bm{\theta}^{(1)} \gets $ A randomly initialized model
            
            \ForAll {$t \in \{1,2,\ldots T\}$}
                \State $\mathcal{C}^{(t)} \gets $ sample a subset of $m$ clients from $\mathcal{C}$
                \State Send the global model $\bm{\theta}^{(t)}$ to every  $c\in \mathcal{C}^{(t)}$
                \State $\bm{\Theta}_t \gets [\bm{\theta}^{(t)}_1, \ldots, \bm{\theta}^{(t)}_m]$ \Comment{Receive the $m$ local models trained by each $c\in \mathcal{C}^{(t)}$}
                \If{$t==1$}
                    \LComment{At the very first round, use the designated fallback aggregation strategy (e.g., FedAvg)}
                    \State $\mathcal{C}^{(t)}_* \gets \texttt{\textsc{Fallback}}(\bm{\Theta}_{t})$
                \Else
                    \LComment{Returns the set of clients classified as legitimate}
                    \State $\mathcal{C}^{(t)}_* \gets \texttt{\textsc{FLANDERS}}(\bm{\Theta}_{t-l:t-1}, \mathcal{C}^{(t)}, k, w, N)$
                \EndIf
                \State $\params^{(t+1)} \gets \phi(\{\params_c^{(t)}~|~c\in \mathcal{C}_*^{(t)}\})$
            \EndFor
        \EndProcedure
    \end{algorithmic}%
\end{algorithm}%

\begin{algorithm}
    \caption{\texttt{\textsc{FLANDERS-Filter}}}\label{alg:flanders}
    \begin{algorithmic}[1]
        \Require The tensor containing the local models of selected clients observed from $l$ past FL rounds ($\bm{\Theta}_{t-l:t-1}$); the clients selected at round $t$ ($\mathcal{C}^{(t)}$); the autoregressive order of MAR ($w$); the number of local models to keep as legitimate ($k$); the number of MAR training iterations ($N$).
        \Ensure The set of $k$ clients classified as legitimate in round $t$ $\mathcal{C}^{(t)}_{*}$.
        \Procedure{\texttt{FLANDERS}}{$\bm{\Theta}_{t-l:t-1}, \mathcal{C}^{(t)}, k, w, N$}
            \LComment{MAR($w$) estimation via ALS training ($N$ iterations)}
            \State $\hat{\bm{\Theta}}_{t} \gets \texttt{MAR}(\bm{\Theta}_{t-l:t-1}, w, N)$
            \State $\bm{s}^{(t)} \gets \delta(\bm{\Theta}_t, \hat{\bm{\Theta}}_t)$ \Comment{Anomaly score vector}
            \State $\mathcal{C}^{(t)}_{*} \gets \{c \in \mathcal{C}^{(t)}~|~s_{c}^{(t)} \leq s_k^{(t)}\}$ \Comment{$s^{(t)}_k$ is the $k$-th smallest anomaly score}
            \State \textbf{return} $\mathcal{C}^{(t)}_{*}$
        \EndProcedure
    \end{algorithmic}
\end{algorithm}

\subsection{Computational Complexity Analysis}
\label{subsec:complexity}
To train our MAR forecasting model, we first need to run the ALS algorithm, which estimates the coefficient matrices $\hat{{\bm A}}$ and $\hat{{\bm B}}$. 
ALS iteratively performs two steps to compute $\hat{{\bm A}}$ and $\hat{{\bm B}}$, respectively. 
At the generic $i$-th iteration, computing the $i$-th $d\times d$ matrix $\hat{{\bm A}}$ costs $O(d^3) + O(dm^2)$; similarly, computing the $i$-th $m\times m$ matrix\footnote{It is worth remarking that, using MAR($1$), $h = m$ and thus $\hat{{\bm B}}$ has size $m\times m$.} $\hat{{\bm B}}$ costs $O(m^3) + O(d^2m)$.
Thus, a single iteration costs $O(d^3) + O(dm^2) + O(m^3) + O(d^2m)$, namely $O(d^3)$ \textit{or} $O(m^3)$,\footnote{$O(d^{2.376})$ \textit{or} $O(m^{2.376})$ using the Coppersmith-Winograd algorithm~\cite{coppersmith1990jsc}.} 
depending what term dominates between $d$ (the number of weights) and $m$ (the number of clients). 
ALS performs the two steps above for $N$ iterations (e.g., in this work, we set $N=100$).
Second, the computation of the anomaly score costs $O(md)$ as it measures the distance between observed and predicted local updates from all the $m$ selected clients.
Third, every time the MAR model is refreshed, we need to rerun ALS; in the worst-case scenario, we might want to re-estimate $\hat{{\bm A}}$ and $\hat{{\bm B}}$ at every single FL round.
\\
It is worth noticing that performing ALS directly on high dimensional parameter space ($d$) in standard FL settings with loads of clients ($m$) may be unfeasible (e.g., when $d$ and $m$ range from $10^6$ to $10^9$).
To keep the computational cost tractable (and limit the impact of the curse of dimensionality), in our experiments, where $d \gg m$, we reduce dimensionality via random sampling on the parameter space, as proposed by~\cite{shejwalkar2021ndss}. Specifically, we sample $\tilde{d} < d$ model parameters for ALS, with $\tilde{d}$ set to $500$.
\section{Experiments}
\label{sec:experiments}

\subsection{Experimental Setup}

\noindent{{\bf {\em Datasets, Tasks, and FL Models.}}}
We consider four public datasets for image classification: {\em MNIST}, {\em Fashion-MNIST}, {\em CIFAR-10}, and {\em CIFAR-100}. 
All datasets are randomly shuffled and partitioned into two disjoint sets: 80\% is used for training and 20\% for testing. 
We train a Multilayer Perceptron (MLP) on \textit{MNIST} and \textit{Fashion-MNIST} and a Convolutional Neural Network (CNN) on \textit{CIFAR-10} and \textit{CIFAR-100} (MobileNet~\cite{howard2017mobilenets}).
All models are trained by minimizing cross-entropy loss. 
\\
The full details are available in Appendix~\ref{app:setup}.
\noindent{{\bf {\em FL Simulation Environment.}}}
To simulate a realistic FL environment, we integrate FLANDERS into Flower \cite{beutel2022flower}. Moreover, we implement in Flower any defense baseline considered that the framework does not natively provide.
In Table~\ref{tab:fl-env} we synthetically report the hyperparameters.
\\
We set the number $K$ of FL clients to $100$, and we assume that the server selects {\em all} these clients in every FL round ($m=K$). In Appendix~\ref{app:client-selection}, we test when the server \textit{randomly} chooses $m<K$ clients. We suppose the attack starts at $t=1$; then, we monitor the performance of the global model until $t=T=50$. 
Recall that we select the malicious clients randomly across all the participating ones; this implies that a single client can alternate between legitimate and malicious behavior over successive FL rounds.
\begin{table}[ht!]
\centering
\caption{Main properties of our FL environment simulated on Flower.
}
\label{tab:fl-env}
\scalebox{1}{
    \begin{tabular}{lc}
    \toprule
    \textbf{Total N. of Clients}        & $K = 100$ \\
    \midrule
    \textbf{N. of Selected Clients (at each round)} & $m = K = 100$\\
    \midrule
    \textbf{Ratio of Malicious Clients} & $r=\{0, 0.2, 0.6, 0.8\}$ \\
    \midrule
    \textbf{Total N. of FL Rounds} & $T=50$ \\ \hline
    \textbf{Autoregressive Order of MAR} & $w=1$ \\
    \midrule
    \textbf{Historical Window Size of Past FL Rounds} & $l=2$ \\
    \midrule
    \textbf{Non-IID Dataset Distribution across Clients} & $\alpha_D = 0.5$  \\
    \bottomrule
    \end{tabular}
}
\end{table}

\noindent{{\bf {\em Non-IID Local Training Data.}}}
We simulate non-iid clients' training data distributions following~\cite{hsu2019LDA}. We assume every client training example is drawn independently with class labels following a categorical distribution over $N$ classes parameterized by a vector $\bm{q}$ such that $q_i \geq 0, i \in [1,\ldots,N]$ and $||\bm{q}||_1 = 1$.
To synthesize a population of non-identical clients, we draw $\bm{q} \sim \text{Dir}(\alpha_D, \bm{p})$ from a Dirichlet distribution, where $\bm{p}$ characterizes a prior class distribution over $N$ classes, and $\alpha_D > 0$ is a {\em concentration} parameter controlling the identicalness among clients. 
With $\alpha_D \rightarrow \infty$, all clients have identical label distributions; on the other extreme, with $\alpha_D \rightarrow 0$, each client holds examples from only one class chosen at random. 
In our experiments, we set $\alpha_D = 0.5$. 

\subsection{Attacks}
\label{subsec:attacks}
We assess the robustness of FLANDERS under the following well-known attacks.
For each attack, we vary the number $b = \lceil r * m\rceil, r\in [0,1]$ of malicious clients, where $r = \{0, 0.2, 0.6, 0.8\}$. Ultimately, we analyze an attacker's performance embedded with a MAR model.

\noindent{{\bf {\em Gaussian Noise Attack.}}} 
\noindent This attack randomly crafts the local models on the compromised clients. 
Specifically, the attacker samples a random value from a Gaussian distribution $\varepsilon \sim \mathcal{N}(0,\sigma^2)$ and sums it to all the $d$ learned parameters. 
We refer to this attack as GAUSS.

\noindent{{\bf {\em ``A Little Is Enough'' Attack}}~\cite{baruch2019neurips}{\bf.}}
This attack 
shifts the aggregation gradient by carefully crafting malicious values that deviate from the correct ones as far as possible. 
We call this attack LIE.

\noindent{{\bf {\em Optimization-based Attack}}~\cite{fang2020usenix}{\bf.}}
This attack is framed as an optimization task, aiming to maximize the distance between the poisoned aggregated gradient and the aggregated gradient under no attack. 
By using a halving search, one can obtain a crafted malicious gradient. 
We refer to this attack as OPT.

\noindent{{\bf {\em AGR Attack Series}}~\cite{shejwalkar2021ndss}{\bf.}}
This improves the optimization program above by introducing perturbation vectors and scaling factors. Then, three instances are proposed: AGR-tailored, AGR-agnostic Min-Max, and Min-Sum, which maximize the deviation between benign and malicious gradients.
In this work, we experiment with AGR Min-Max, which we call AGR-MM.

Below, we describe the critical parameters for
each attack considered, which are also summarized in Table~\ref{tab:attacks}.

\noindent{{\bf {\em GAUSS}}{\bf.}} This attack has only one parameter: the magnitude $\sigma$ of the perturbation to apply. We set $\sigma=10$ for all the experiments.

\noindent{{\bf {\em LIE}}{\bf.}} This method has no parameters to set.

\noindent{{\bf {\em OPT}}{\bf.}} The parameter $\tau$ represents the minimum value that $\lambda$ can assume. Below this threshold, the halving search stops. As suggested by the authors, we set $\tau=10^{-5}$.

\noindent{{\bf {\em AGR-MM}}{\bf.}} In addition to the threshold $\tau$, AGR-MM uses the perturbation vectors $\nabla^p$ in combination with the scaling coefficient $\gamma$ to optimize. We set $\tau=10^{-5}$ and $\nabla^p = \{std\}$, which is the vector obtained by computing the parameters' inverse of the standard deviation. For Krum, we set $\nabla^p = \{uv\}$, which is the inverse unit vector perturbation.
\begin{table}[ht]
\centering
\caption{Key parameter settings for each attack strategy considered.
}
\label{tab:attacks}
\begin{tabular}{cc}
    \toprule
    \textbf{Attack}        & \textbf{Parameters} \\
    \midrule
    {\bf {\em GAUSS}} & $\sigma=10$\\
    \midrule
    {\bf {\em LIE}} & N/A\\
    \midrule
    {\bf {\em OPT}} & $\tau=10^{-5}$\\
    \midrule
    {\bf {\em AGR-MM}} & $\tau=10^{-5}; \nabla^p = \{uv, std\}; \gamma=5$\\
    \bottomrule
\end{tabular}
\end{table}

\subsection{Evaluation}
\label{subsec:eval}
We evaluate four key aspects of FLANDERS. 
Firstly, we test its ability to detect malicious clients against the best-competing filtering strategy, FLDetector.
Secondly, we measure the accuracy improvement of the global model when FLANDERS is paired with ``vanilla'' FedAvg and the most popular robust aggregation baselines: FedMedian, Trimmed Mean, Multi-Krum, Bulyan, DnC. 
Thirdly, we analyze the cost-benefit trade-off of FLANDERS.
Lastly, we test the robustness of our method against adaptive attacks. 

\smallskip
\noindent{\textbf{\textit{Malicious Detection Accuracy}.}} \label{subsec:detect-acc}
Table~\ref{tab:pr-20} shows the precision ($P$) and recall ($R$) of FLDetector and FLANDERS in filtering out malicious clients across different datasets and attacks, with $r=0.2$ (20\% evil participants in the FL systems). 
Remarkably, FLANDERS successfully detects \textit{all and only} malicious clients across every attack setting except for OPT, outperforming its main competitor, FLDetector.
This is further confirmed by Table~\ref{tab:increment-fld}, which shows that our method generally provides much higher protection than FLDetector when combined with standard FedAvg.

Better results are observed for extreme attack settings ($r=0.8$) in Tables~\ref{tab:pr-80}, and \ref{tab:increment-fld-80}, where OPT maximizes the distance between the legitimate models and the malicious ones by erroneously thinking that the high number of its controlled clients pierces through the aggregation functions. Results for $r=0.6$ in  Appendix~\ref{app:flanders-accuracy} are consistent with these findings.

FLDetector fails to distinguish malicious updates from legitimate ones because \textit{(i)} the anomaly score is based on approximating a single Hessian matrix that must be close to all legitimate clients, making it inadequate for highly non-iid settings, where malicious updates can be similar to legitimate ones; and \textit{(ii)} the suspicious scores are computed as the average normalized Euclidean distance of the past iterations. Although this is coherent and works well with a threat model where the malicious clients are always the same across all rounds, FLANDERS does not make such an assumption, making FLDetector unable to recognize malicious updates arriving for the first time from an ex-legitimate client.

\begin{table*}[htb!]
\centering
\vspace{-4mm}
\caption{\textit{Precision} ($P$) and \textit{Recall} ($R$) of FLDetector and FLANDERS in detecting malicious clients across various datasets and attacks ($r=0.2$; $T=50$ rounds).} 
\label{tab:pr-20}
\scalebox{1}{
\begin{tabular}{@{}cccccccccccccccccc@{}}\toprule
 \multicolumn{1}{c}{} & \multicolumn{8}{c}{\textbf{FLDetector}} & \phantom{a} & \multicolumn{8}{c}{\textbf{FLANDERS}} \\
 \cmidrule(lr){2-9} \cmidrule(lr){11-18}
 & \multicolumn{2}{c}{GAUSS} & \multicolumn{2}{c}{LIE} & \multicolumn{2}{c}{OPT} & \multicolumn{2}{c}{AGR-MM} && \multicolumn{2}{c}{GAUSS} & \multicolumn{2}{c}{LIE} & \multicolumn{2}{c}{OPT} & \multicolumn{2}{c}{AGR-MM} \\
 \cmidrule(lr){2-3} \cmidrule(lr){4-5} \cmidrule(lr){6-7} \cmidrule(lr){8-9} \cmidrule(lr){11-12} \cmidrule(lr){13-14} \cmidrule(lr){15-16} \cmidrule(lr){17-18}
                            & $P$   & $R$   & $P$   & $R$   & $P$    & $R$    & $P$   & $R$     && $P$   & $R$   & $P$   & $R$    & $P$  & $R$   & $P$  & $R$ \\
\midrule

\textit{MNIST}          & $0.20$ & $0.20$ & $0.22$ & $0.22$ & ${\bf 0.20}$ & ${\bf 0.20}$ & $0.19$ & $0.19$  && ${\bf 1.0}$ & ${\bf 1.0}$ & ${\bf 1.0}$ & ${\bf 1.0}$ & $0.13$ & $0.13$ & ${\bf 1.0}$ & ${\bf 1.0}$ \\

\textit{Fashion-MNIST}  & $0.21$ & $0.21$ & $0.18$ & $0.18$ & ${\bf 0.21}$ & ${\bf 0.21}$ & $0.20$ & $0.20$  && ${\bf 1.0}$ & ${\bf 1.0}$ & ${\bf 1.0}$ & ${\bf 1.0}$ & $0.13$ & $0.13$ & ${\bf 1.0}$ & ${\bf 1.0}$ \\

\textit{CIFAR-10}       & $0.21$ & $0.21$ & $0.21$ & $0.21$ & $0.22$ & $0.22$ & $0.19$ & $0.19$  && ${\bf 1.0}$ & ${\bf 1.0}$ & ${\bf 1.0}$ & ${\bf 1.0}$ & ${\bf 0.58}$ & ${\bf 0.58}$ & ${\bf 1.0}$ & ${\bf 1.0}$ \\

\textit{CIFAR-100}      & $0.20$ & $0.20$ & $0.19$ & $0.19$ & $0.20$ & $0.20$ & $0.21$ & $0.21$  && ${\bf 1.0}$ & ${\bf 1.0}$ & ${\bf 1.0}$ & ${\bf 1.0}$ & ${\bf 1.0}$  & ${\bf 1.0}$  & ${\bf 1.0}$ & ${\bf 1.0}$ \\
\bottomrule
\end{tabular}
}
\end{table*}

\begin{table*}[htb!]
\centering
\caption{Accuracy of the global model using FedAvg with FLDetector and FLANDERS ($r=0.2$).} 
\label{tab:increment-fld}
\scalebox{0.95}{
\begin{tabular}{ccccccccccccccc}
\toprule
& \multicolumn{4}{c}{\textbf{FedAvg}} & \phantom{a} & \multicolumn{4}{c}{\textbf{FLDetector + FedAvg}} & \phantom{a} & \multicolumn{4}{c}{\textbf{FLANDERS + FedAvg}} \\
\cmidrule(lr){2-5} \cmidrule(lr){7-10} \cmidrule(lr){12-15}
                    &  GAUSS    &  LIE     &  OPT     &  AGR-MM  &&  GAUSS   &  LIE     &  OPT     &  AGR-MM  &&  GAUSS    &  LIE     &  OPT     &  AGR-MM      \\ 
\midrule
\textit{MNIST}          & $0.18$ & $0.12$ & $0.63$ & $0.34$ && $0.20$ & $0.11$ & ${\bf 0.68}$ & $0.43$ && ${\bf 0.86}$ & ${\bf 0.83}$ & $0.62$       & ${\bf 0.85}$ \\
\textit{Fashion-MNIST}  & $0.25$ & $0.10$ & $0.56$ & $0.16$ && $0.28$ & $0.10$ & ${\bf 0.60}$ & $0.17$ && ${\bf 0.69}$ & ${\bf 0.64}$ & $0.58$       & ${\bf 0.63}$ \\
\textit{CIFAR-10}       & $0.10$ & $0.10$ & $0.23$ & $0.10$ && $0.10$ & $0.10$ & $0.26$       & $0.12$ && ${\bf 0.38}$ & ${\bf 0.37}$ & ${\bf 0.28}$ & ${\bf 0.36}$ \\
\textit{CIFAR-100}      & $0.01$ & $0.01$ & $0.01$ & $0.02$ && $0.01$ & $0.01$ & $0.01$       & $0.02$ && ${\bf 0.07}$ & ${\bf 0.05}$ & ${\bf 0.05}$ & ${\bf 0.06}$ \\
\bottomrule
\end{tabular}
}
\end{table*}

\begin{table*}[htb!]
\centering
\caption{\textit{Precision} ($P$) and \textit{Recall} ($R$) of FLDetector and FLANDERS in detecting malicious clients across various datasets and attacks ($r=0.8$; $T=50$ rounds).}
\label{tab:pr-80}
\scalebox{1}{
\begin{tabular}{@{}cccccccccccccccccc@{}}\toprule
 \multicolumn{1}{c}{} & \multicolumn{8}{c}{\textbf{FLDetector}} & \phantom{a} & \multicolumn{8}{c}{\textbf{FLANDERS}} \\
 \cmidrule(lr){2-9} \cmidrule(lr){11-18}
 & \multicolumn{2}{c}{GAUSS} & \multicolumn{2}{c}{LIE} & \multicolumn{2}{c}{OPT} & \multicolumn{2}{c}{AGR-MM} && \multicolumn{2}{c}{GAUSS} & \multicolumn{2}{c}{LIE} & \multicolumn{2}{c}{OPT} & \multicolumn{2}{c}{AGR-MM} \\
 \cmidrule(lr){2-3} \cmidrule(lr){4-5} \cmidrule(lr){6-7} \cmidrule(lr){8-9} \cmidrule(lr){11-12} \cmidrule(lr){13-14} \cmidrule(lr){15-16} \cmidrule(lr){17-18}
                            & $P$   & $R$   & $P$   & $R$   & $P$    & $R$    & $P$   & $R$     && $P$   & $R$   & $P$   & $R$    & $P$  & $R$   & $P$  & $R$ \\
\midrule
\textit{MNIST}              & $0.80$ & $0.80$ & $0.80$ & $0.80$ & $0.80$ & $0.80$ & $0.80$ & $0.80$ && ${\bf 1.0}$ & ${\bf 1.0}$ & ${\bf 1.0}$ & ${\bf 1.0}$ & ${\bf 1.0}$ & ${\bf 1.0}$ & ${\bf 1.0}$ & ${\bf 1.0}$ \\
\textit{Fashion-MNIST}      & $0.80$ & $0.80$ & $0.80$ & $0.80$ & $0.79$ & $0.79$ & $0.80$ & $0.80$ && ${\bf 1.0}$ & ${\bf 1.0}$ & ${\bf 1.0}$ & ${\bf 1.0}$ & ${\bf 1.0}$ & ${\bf 1.0}$ & ${\bf 1.0}$ & ${\bf 1.0}$ \\
\textit{CIFAR-10}           & $0.80$ & $0.80$ & $0.80$ & $0.80$ & $0.80$ & $0.80$ & $0.80$ & $0.80$ && ${\bf 1.0}$ & ${\bf 1.0}$ & ${\bf 1.0}$ & ${\bf 1.0}$ & ${\bf 1.0}$ & ${\bf 1.0}$ & ${\bf 1.0}$ & ${\bf 1.0}$ \\
\textit{CIFAR-100}          & $0.80$ & $0.80$ & $0.80$ & $0.80$ & $0.80$ & $0.80$ & $0.80$ & $0.80$ && ${\bf 1.0}$ & ${\bf 1.0}$ & ${\bf 1.0}$ & ${\bf 1.0}$ & ${\bf 1.0}$  & ${\bf 1.0}$  & ${\bf 1.0}$ & ${\bf 1.0}$ \\
\bottomrule
\end{tabular}
}
\end{table*}

\begin{table*}[htb!]
\centering
\caption{Accuracy of the global model using FedAvg with FLDetector and FLANDERS ($r=0.8$).}
\label{tab:increment-fld-80}
\scalebox{0.95}{
\begin{tabular}{ccccccccccccccc}
\toprule
& \multicolumn{4}{c}{\textbf{FedAvg}} & \phantom{a} & \multicolumn{4}{c}{\textbf{FLDetector + FedAvg}} & \phantom{a} & \multicolumn{4}{c}{\textbf{FLANDERS + FedAvg}} \\
\cmidrule(lr){2-5} \cmidrule(lr){7-10} \cmidrule(lr){12-15}
                    &  GAUSS    &  LIE     &  OPT     &  AGR-MM  &&  GAUSS   &  LIE     &  OPT     &  AGR-MM  &&  GAUSS    &  LIE     &  OPT     &  AGR-MM      \\ 
\midrule
\textit{MNIST}          & $0.18$ & $0.11$ & $0.21$ & $0.11$ && $0.15$ & $0.13$ & $0.23$ & $0.12$ && ${\bf 0.75}$ & ${\bf 0.84}$ & ${\bf 0.84}$ & ${\bf 0.82}$ \\
\textit{Fashion-MNIST}  & $0.24$ & $0.10$ & $0.19$ & $0.10$ && $0.19$ & $0.10$ & $0.03$ & $0.10$ && ${\bf 0.68}$ & ${\bf 0.70}$ & ${\bf 0.66}$ & ${\bf 0.66}$ \\
\textit{CIFAR-10}       & $0.10$ & $0.10$ & $0.11$ & $0.10$ && $0.10$ & $0.10$ & $0.10$ & $0.10$ && ${\bf 0.33}$ & ${\bf 0.32}$ & ${\bf 0.32}$ & ${\bf 0.32}$ \\
\textit{CIFAR-100}      & $0.01$ & $0.01$ & $0.01$ & $0.01$ && $0.01$ & $0.01$ & $0.01$ & $0.01$ && ${\bf 0.09}$ & ${\bf 0.11}$ & ${\bf 0.11}$ & ${\bf 0.10}$ \\
\bottomrule
\end{tabular}
}
\end{table*}


\smallskip
\noindent{\textbf{\textit{Aggregation Robustness Lift}.}} \label{subsec:agg-robustness-lift}
We proceed to evaluate the enhancement that FLANDERS provides to the robustness of the global model. 
Specifically, we measure the best accuracy of the global model under several attack strengths using all the baselines \textit{without} and \textit{with} FLANDERS as a pre-filtering strategy.
Table~\ref{tab:increment} shows that FLANDERS keeps high accuracy for the best global model under extreme attacks ($r=0.8$) when used before \textit{every} aggregation method, including Multi-Krum and Bulyan, which would otherwise be inapplicable in such strong attack scenarios.
This is further emphasized in Table~\ref{tab:robustness-1} (see Appendix~\ref{app:agg-lift}), where Multi-Krum, when paired with FLANDERS, can effectively operate without \textit{any} performance degradation, even in the presence of $80\%$ malicious clients.

\begin{figure}[htb!]
     \centering
     \includegraphics[width=\columnwidth]{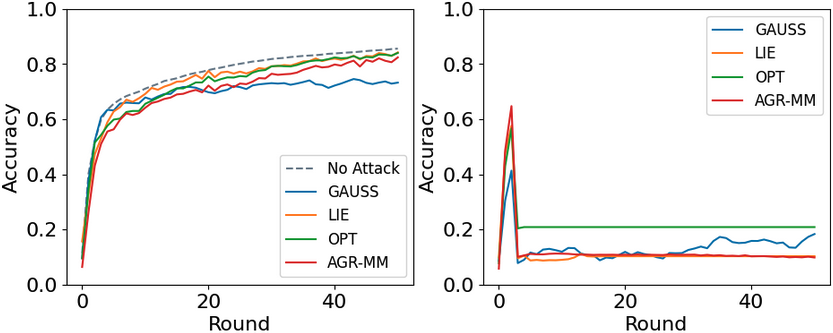}
     \caption{FedAvg with FLANDERS (left) vs. "vanilla" FedAvg (right). Accuracy of the global model in each FL round under all attack strategies on the \textit{MNIST} dataset, with $80\%$ of malicious clients. Attack starts at round $t=3$.}
     \label{fig:accuracy-over-time}
\end{figure}

In Figure~\ref{fig:accuracy-over-time} we compare the accuracy of the global model when using FLANDERS (left) and when using FedAvg (right) across the whole FL training process. The left figure shows how the evolution of the accuracy over multiple rounds remains stable and similar to the one without any attack (dashed line), while on the right the accuracy drops irremediably.

\begin{table*}[htb!]
\centering
\caption{Accuracy of the global model using all the baseline aggregations without and with FLANDERS ($r=0.8$).}
\label{tab:increment}
\scalebox{0.75}{
\begin{tabular}{cccccccccccccccccccc}
\toprule
& \multicolumn{4}{c}{\textit{MNIST}} & \phantom{a} & \multicolumn{4}{c}{\textit{Fashion-MNIST}} & \phantom{a} & \multicolumn{4}{c}{\textit{CIFAR-10}} & \phantom{a} & \multicolumn{4}{c}{\textit{CIFAR-100}} \\
\cmidrule(lr){2-5} \cmidrule(lr){7-10} \cmidrule(lr){12-15} \cmidrule(lr){17-20}
              &  GAUSS        &  LIE          &  OPT          &  AGR-MM       &&  GAUSS        &  LIE          &  OPT          &  AGR-MM       &&  GAUSS         &  LIE          &  OPT          &  AGR-MM        &&  GAUSS         &  LIE          &  OPT          &  AGR-MM      \\ 
\midrule
FedAvg      & $0.18$        & $0.11$        & $0.21$        & $0.11$        && $0.24$       & $0.10$        & $0.19$        & $0.10$        && $0.10$       & $0.10$        & $0.11$        & $0.10$        && $0.01$       & $0.01$        & $0.01$        & $0.01$  \\
+ FLANDERS  & ${\bf 0.75}$  & ${\bf 0.84}$  & ${\bf 0.84}$  & ${\bf 0.82}$  && ${\bf 0.68}$ & ${\bf 0.70}$  & ${\bf 0.66}$  & ${\bf 0.66}$  && ${\bf 0.33}$ & ${\bf 0.32}$  & ${\bf 0.32}$  & ${\bf 0.32}$  && ${\bf 0.09}$ & ${\bf 0.11}$  & ${\bf 0.11}$  & ${\bf 0.10}$  \\
\midrule
FedMedian   & $0.34$ & $0.19$ & $0.13$ & $0.23$ && $0.29$ & $0.10$ & $0.17$ & $0.10$ && $0.13$ & $0.10$ & $0.10$ & $0.12$ && $0.01$ & $0.01$ & $0.01$ & $0.01$  \\
+ FLANDERS  & ${\bf 0.81}$ & ${\bf 0.84}$ & ${\bf 0.85}$ & ${\bf 0.81}$ && ${\bf 0.70}$ & ${\bf 0.71}$ & ${\bf 0.71}$ & ${\bf 0.68}$ && ${\bf 0.29}$ & ${\bf 0.29}$ & ${\bf 0.31}$ & ${\bf 0.28}$ && ${\bf 0.10}$ & ${\bf 0.11}$ & ${\bf 0.11}$ & ${\bf 0.10}$  \\
\midrule
TrimmedMean & $0.15$ & $0.13$ & $0.20$ & $0.18$ && $0.21$ & $0.10$ & $0.23$ & $0.10$ && $0.10$ & $0.10$ & $0.10$ & $0.10$ && $0.01$ & $0.01$ & $0.01$ & $0.01$ \\
+ FLANDERS  & ${\bf 0.81}$ & ${\bf 0.83}$ & ${\bf 0.83}$ & ${\bf 0.85}$ && ${\bf 0.71}$ & ${\bf 0.70}$ & ${\bf 0.71}$ & ${\bf 0.69}$ && ${\bf 0.30}$ & ${\bf 0.29}$ & ${\bf 0.30}$ & ${\bf 0.29}$ && ${\bf 0.11}$ & ${\bf 0.10}$ & ${\bf 0.11}$ & ${\bf 0.11}$ \\
\midrule
Multi-Krum  & $0.80$ & $0.12$ & $0.17$ & $0.25$ && $0.64$ & $0.10$ & $0.20$ & $0.10$ && $0.34$ & $0.10$ & $0.10$ & $0.10$ && $0.08$ & $0.01$ & $0.01$ & $0.01$ \\
+ FLANDERS  & ${\bf 0.87}$ & ${\bf 0.90}$ & ${\bf 0.88}$ & ${\bf 0.89}$ && ${\bf 0.69}$ & ${\bf 0.68}$ & ${\bf 0.72}$ & ${\bf 0.68}$ && ${\bf 0.38}$ & ${\bf 0.38}$ & ${\bf 0.39}$ & ${\bf 0.40}$ && ${\bf 0.11}$ & ${\bf 0.10}$ & ${\bf 0.10}$ & ${\bf 0.11}$ \\
\midrule
Bulyan      & N/A & N/A & N/A & N/A && N/A & N/A & N/A & N/A && N/A & N/A & N/A & N/A && N/A & N/A & N/A & N/A \\
+ FLANDERS  & ${\bf 0.89}$ & ${\bf 0.85}$ & ${\bf 0.88}$ & ${\bf 0.82}$ && ${\bf 0.68}$ & ${\bf 0.66}$ & ${\bf 0.68}$ & ${\bf 0.70}$ && ${\bf 0.40}$ & ${\bf 0.43}$ & ${\bf 0.40}$ & ${\bf 0.41}$ && ${\bf 0.11}$ & ${\bf 0.11}$ & ${\bf 0.11}$ & ${\bf 0.11}$ \\
\midrule
DnC         & $0.21$ & $0.11$ & $0.17$ & $0.11$ && $0.25$ & $0.10$ & $0.14$ & $0.10$ && $0.10$ &$ 0.10$ & $0.10$ & $0.10$ && $0.01$ & $0.01$ & $0.01$ & $0.01$ \\
+ FLANDERS  & ${\bf 0.85}$ & ${\bf 0.87}$ & ${\bf 0.89}$ & ${\bf 0.87}$ && ${\bf 0.71}$ & ${\bf 0.69}$ & ${\bf 0.68}$ & ${\bf 0.68}$ && ${\bf 0.41}$ & ${\bf 0.40}$ & ${\bf 0.39}$ & ${\bf 0.40}$ && ${\bf 0.10}$ & ${\bf 0.11}$ & ${\bf 0.11}$ & ${\bf 0.12}$ \\
\bottomrule
\end{tabular}
}
\end{table*}

For weaker attacks, e.g., $r=0.2$, FLANDERS still generally improves the accuracy of the global model, except when combined with Bulyan, which alone appears already robust enough to counter these mild attacks (see Table~\ref{tab:increment-20} in Appendix~\ref{app:agg-lift}). The complete results are in Appendix~\ref{app:agg-lift}.
However, it is worth remarking that current robust aggregation methods like Multi-Krum and Bulyan are not designed to handle stronger attack settings. 
Nevertheless, when these methods are combined with FLANDERS, they can be deployed successfully even under extremely severe attacks due to FLANDERS' capability to filter out every malicious client \textit{before} the aggregation process takes place.
This is illustrated with an example in Table~\ref{tab:robustness-1}, where it is demonstrated that Multi-Krum, when paired with FLANDERS, can effectively operate without \textit{any} performance degradation even when the proportion of malicious clients reaches $80\%$. 
Similarly, as shown in Fig.~\ref{fig:accuracy-over-time}, the performance remains robust across all FL rounds even when FLANDERS is paired with a simple aggregation function, such as FedAvg. Instead, when our filter is removed, the accuracy drops as soon as the attack begins.

\begin{figure}[htb!]
     \centering
     \includegraphics[width=\columnwidth]{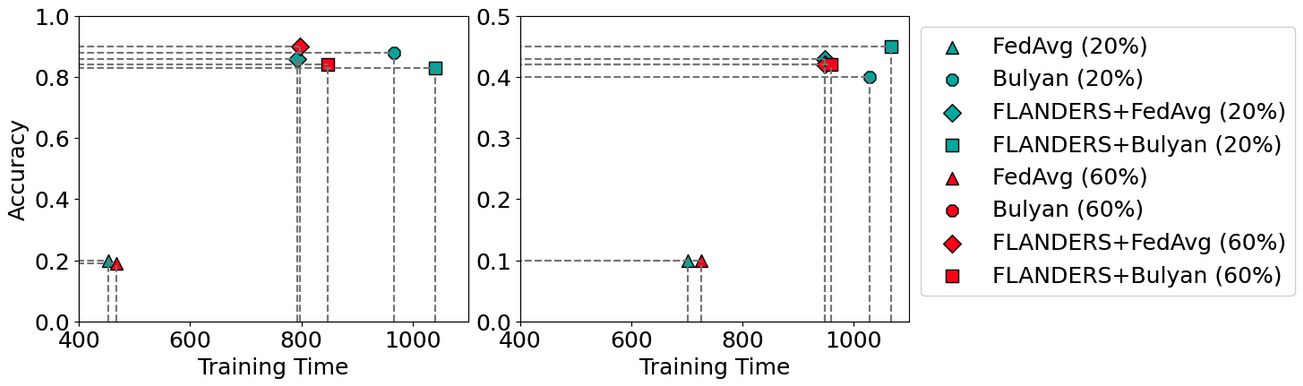}
     \caption{Accuracy vs. \textit{total} Training Time (in seconds) of FedAvg and Bulyan compared with their corresponding versions with FLANDERS as a filter for the \textit{MNIST} (left) and \textit{CIFAR-10} (right) datasets.}
     \label{fig:tradeoff}
\end{figure}

\smallskip
\noindent{\textbf{\textit{Cost-Benefit Analysis}.}} \label{subsec:cost-benefit}
Obviously, the robustness guaranteed by FLANDERS under extreme attack scenarios comes with costs, especially due to the MAR estimation stage. 
Fig.~\ref{fig:tradeoff} depicts two scatter plots for the \textit{MNIST} and \textit{CIFAR-10} datasets, focusing on a specific attack scenario (AGR-MM). Each data point on a scatter plot represents a method under one of two attack strengths considered ($r=0.2$ and $r=0.6$). These data points are specified by two coordinates: the overall training time on the $x$-axis and the maximum accuracy of the global model obtained after $T=50$ FL rounds on the $y$-axis.

Overall, the take-home message is as follows. In scenarios with low attack strength ($r=0.2$), Bulyan demonstrates superior accuracy, while FLANDERS + FedAvg offers comparable performance with notably shorter training times. However, as the attack strength increases ($r=0.6$), Bulyan becomes impractical, FedAvg alone proves ineffective, and FLANDERS emerges as the optimal choice for achieving the best accuracy vs. cost trade-off.

\smallskip
\noindent{\textbf{\textit{Robustness against Adaptive Attacks}.}} \label{subsec:adaptive-attack}
In this section, we further validate the robustness of FLANDERS against adaptive attacks. We consider a scenario where malicious clients are aware that the FL server uses our method as a pre-aggregation filter. 
Specifically, we focus on two levels of knowledge. The first scenario assumes that malicious clients tentatively guess the subset of parameters ($\widetilde{d}$) used by the FL server to estimate the MAR forecasting model. We refer to this setting as \textit{non-omniscient}.
The second, more challenging as well as unrealistic scenario assumes that malicious clients know \textit{exactly} which parameters are used by the FL server. We call this second scenario \textit{omniscient}. Obviously, the latter penalizes FLANDERS way more than the former. 
\\
Specifically, we perform our experiments over $T=20$ rounds with $m=20$ clients, of which $r=0.2$ ($b=5$) are malicious. The attacker constructs a matrix $M = b \times \widetilde{d}$ using the local models generated by the corrupted clients. This matrix $M$ is then passed as input to the same forecasting model, MAR, that the server uses to determine the legitimacy of local models. The attacker, instead, substitutes the legitimate parameters with those estimated by MAR, exploiting the fact that these estimations do not perform like a legitimate local model. This substitution ultimately hurts the accuracy of the global model once the parameters are aggregated. 
\\
As introduced above, we first assume that the attacker is \textit{non-omniscient}, meaning it does not know which parameters the server has selected for the MAR estimation. Instead, the attacker selects the last layer as $\widetilde{d}$. 
Afterward, we consider an \textit{omniscient} attacker who exploits the knowledge of the parameters selected by the server. 
In Table~\ref{tab:nonomn-adaptive}, we show the results of the non-omniscient scenario, where FLANDERS + Multi-Krum outperforms all other baselines on all three datasets. Table~\ref{tab:omn-adaptive}, on the other hand, refers to the omniscient scenario and demonstrates a different pattern, where FedAvg and Multi-Krum alone perform better than when coupled with FLANDERS.
This may be because FLANDERS consistently filters out legitimate local models in favor of corrupted ones. When using FedAvg, the impact of corrupted parameters is mitigated by averaging a larger number of legitimate models and because the corrupted models' parameter values are not too different, unlike in methods like OPT. On the other hand, Multi-Krum selects parameters with more nearby neighbors, and with only $b=5$ corrupted clients, legitimate models likely still have more and closer neighbors, effectively defending against our adaptive attack.

\begin{table}[htb!]
\centering
\caption{Accuracy of the global model using FedAvg and Multi-Krum, with and without FLANDERS, under the \textit{\textbf{non-}omniscient} adaptive attack.}
\label{tab:nonomn-adaptive}
\begin{tabular}{cccc}
    \toprule
    Strategy & \textit{MNIST} & \textit{Fashion-MNIST} & \textit{CIFAR-10} \\
    \midrule
    FedAvg                  & $0.84$ & $0.68$ & $0.45$ \\
    FLANDERS + FedAvg       & $0.82$ & $0.67$ & $0.43$ \\
    Multi-Krum              & $0.87$ & ${\bf 0.72}$ & $0.41$ \\
    FLANDERS + Multi-Krum   & ${\bf 0.90}$ & ${\bf 0.72}$ & ${\bf 0.47}$ \\
    \bottomrule
\end{tabular}
\end{table}

\begin{table}[htb!]
\centering
\caption{Accuracy of the global model using FedAvg and Multi-Krum, with and without FLANDERS, under the \textit{omniscient} adaptive attack.}
\label{tab:omn-adaptive}
\begin{tabular}{cccc}
    \toprule
    Strategy & \textit{MNIST} & \textit{Fashion-MNIST} & \textit{CIFAR-10} \\
    \midrule
    FedAvg                  & $0.78$ & ${\bf 0.68}$ & ${\bf 0.25}$ \\
    FLANDERS + FedAvg       & $0.65$ & $0.61$ & $0.10$ \\
    Multi-Krum              & ${\bf 0.86}$ & ${\bf 0.68}$ & $0.23$ \\
    FLANDERS + Multi-Krum   & $0.73$ & $0.60$ & $0.10$ \\
    \bottomrule
\end{tabular}
\end{table}

\section{Limitations and Future Work}
\label{sec:limitations}

\subsection{Efficiency/Feasibility}
\label{subsec:efficiency}
As we discussed in Section~5.3 and Appendix~\ref{app:defenses}, FLANDERS may suffer from a high computational cost that could limit its deployment in practice.
This concern holds particularly true for \textit{cross-device} FL configurations encompassing millions of edge devices. Conversely, the impact on \textit{cross-silo} FL scenarios would be notably less pronounced. 
However, as we have introduced FLANDERS as a versatile and robust aggregation approach applicable to diverse FL setups (cross-silo and cross-device), there are implementation techniques available to mitigate its complexity. For instance, methods like random parameter sampling~\cite{shejwalkar2021ndss} can be employed, and we have already incorporated them.
Still, we plan to enhance the scalability of FLANDERS further in future work. 
For example, we could replace the standard matrix inversion algorithm with the more efficient Coppersmith-Winograd algorithm~\cite{coppersmith1990jsc} and find the optimal frequency for re-estimating FLANDERS' parameters, instead of performing it during each FL round.

In an extended version of this work, we plan to include a parameter sensitivity analysis, where the performance of FLANDERS is evaluated based on the number $d$ of sampled parameters. This analysis will hopefully provide insights into the optimal trade-off between robustness and efficiency. Additionally, more sophisticated parameter selection strategies beyond pure random sampling could be explored to focus on the most informative neurons, such as Neuron Shapley~\cite{neuron_shapley}.

\subsection{Potential Privacy Leakage}
\label{sec:limitations-privacy}
In the standard FL setup, the central server must access the local model updates sent by each client (e.g., even to perform a simple FedAvg). 
Therefore, our approach, FLANDERS, does not need additional permission nor violate any privacy constraints beyond what any other FL server could already do. 
Indeed, the most effective robust aggregation schemes, such as Krum and Bulyan, like FLANDERS, assume that the central server is a \textit{trusted} entity. 
However, if this assumption does not hold, the scenario will change. For instance, if the server operates as an ``honest but curious'' entity, thoroughly examining the local model parameters received for training the outlier detection model (such as MAR) could unveil sensitive details that the server might exploit, potentially inferring information about each client's local data distribution.

\subsection{Cross-Device Setting}
The cross-device setting (thousands to millions of clients) penalizes FLANDERS, as the server cannot select them all, and the probability of choosing the same client in consecutive rounds is low. However, FLANDERS is more appropriate in cross-silo FL (tens to hundreds of clients) than in cross-device settings. In fact, large attacks involving more than 50\% of clients are less feasible when the total number of participants grows to the order of thousands \cite{shejwalkar2022sp}. In such cases, since selected clients have little or no history, FLANDERS computes distances between local updates and the last global model, turning it into a heuristic similar to (Multi-)Krum. Thus, FLANDERS will be comparable to any other heuristic.

\subsection{Benchmarking}
We extensively validate FLANDERS with an exhaustive set of experiments. We compared it against six robust baselines amongst the most powerful at the time of writing, along with standard FedAvg.
\\
In addition, robust federated aggregation is a hot research topic; keeping pace with the massive body of work that has been flourishing is challenging.
Hence, we might have missed considering some other methods. 
However, we believe that the value of our work still stands. In this regard, our contribution is clear: We are the first to frame the problem of detecting untargeted model poisoning attacks on FL as a matrix-valued time series anomaly detection task and to propose a method effective under severe attack settings, as opposed to existing baselines. 
\section{Conclusion}
\label{sec:conclusion}
We introduced FLANDERS, a novel FL filter robust to extreme untargeted model poisoning attacks, i.e., when malicious clients far exceed legitimate participants. 
FLANDERS serves as a pre-processing step before applying aggregation rules, enhancing robustness across diverse hostile settings.
\\
FLANDERS treats the sequence of local model updates sent by clients in each FL round as a matrix-valued time series. 
Then, it identifies malicious client updates as outliers in this time series using a matrix autoregressive forecasting model.
\\
Experiments conducted in several non-iid FL setups demonstrated that existing (secure) aggregation methods further improve their robustness when paired with FLANDERS. 
Moreover, FLANDERS allows these methods to operate even under extremely severe attack scenarios thanks to its ability to accurately filter out every malicious client \textit{before} the aggregation process takes place.
\\
In the future, we will address the primary limitations of this work, as discussed in Section~\ref{sec:limitations}.

\bibliographystyle{plain}
\bibliography{references}

\clearpage
\section*{Appendix}
\label{sec:appendix}
The (anonymous) GitHub repository with the code and the data to replicate the results discussed in this work is accessible at the following link: \\\url{https://anonymous.4open.science/r/flanders_exp-7EEB}

\section{A Note on the Terminology Used}
\label{app:terminology}
The type of model poisoning attacks we consider here are often referred to as \textit{Byzantine} attacks in the literature (\cite{blanchard2017nips,fang2020usenix,barroso2023if}). 
Although, in this work, we adhere to the taxonomy proposed by~\cite{barroso2023if}, the research community has yet to reach a unanimous consensus on the terminology. 
In fact, some authors use the word ``Byzantine'' as an umbrella term to broadly indicate \textit{any} attack involving malicious clients (e.g., targeted data poisoning like backdoor attacks \textit{and} untargeted model poisoning as in~\cite{hu2021arxiv}).
Therefore, to avoid confusion and hurting the feelings of some readers who have already debated on that and found the term inappropriate or disrespectful,\footnote{\small{\url{https://openreview.net/forum?id=pfuqQQCB34&noteId=5KAMwoI2cC}}} we have decided \textit{not} to use the word ``Byzantine'' to refer to our attack model.

\section{The Impact of Malicious Clients at each FL Round}
\label{app:impact}
Under our assumptions, the FL system contains $K$ clients, where $b$ of them are malicious and controlled by an attacker ($0\leq b\leq K$).
In addition, at each FL round, $m$ clients ($1 \leq m \leq K$) are selected, and thus, some of the $m$ model updates received by the central server may be corrupted.
The probability of this event can actually be computed by noticing that the outcome of the client selection at each round can be represented by a random variable $X\sim \text{Hypergeometric}(K, b, m)$, whose probability mass function is:
\[p_X(x) = \Pr(X=x) = \frac{\binom{b}{x} \binom{K-b}{m-x}}{\binom{K}{m}}.
\]
The chance that, at a single round, {\em at least one} of the $b$ malicious clients ends up in the list of $m$ clients randomly picked by the server is equal to:
\[\Pr(X\geq 1) = 1 - \Pr(X=0) = 1 - \frac{\binom{b}{0}\binom{K-b}{m-0}}{\binom{K}{m}}= 1 - \frac{ \binom{K-b}{m}}{\binom{K}{m}}.
\]
For example, if the total number of clients is $K=100$, $b=5$ of them are malicious, and $m=20$ must be drawn at each round, then $\Pr(X\geq 1) \approx 68\%$. 
In other words, there are about two out of three chances that at least one malicious client is selected at {\em every} FL round.

In our FL simulation environment, Flower, we can set a fixed proportion of malicious clients in the system (e.g., $20\%$). However, it is important to note that these clients may not remain constant across different FL rounds. In other words, a client who is selected in one round and acts legitimately could become malicious in another round of the FL process.

\section{Matrix Autoregressive Model (MAR)}
\label{app:mar}
This work assumes the temporal evolution of the local models sent by clients at each FL round exhibits a bilinear structure captured by a \textit{matrix autoregressive model} of order $1$, i.e., a Markovian forecasting model denoted by MAR($1$) and defined as follows:
\[
{\bm \Params}_{t} = {\bm A}{\bm \Params}_{t-1}{\bm B} + {\bm E}_t,
\]
where ${\bm E}_t$ is a white noise matrix, i.e., its entries are iid normal with zero-mean and constant variance.
To approximate such a behavior, we consider a parametric forecasting model $f$, in the form:
\[
\widetilde{\bm{\Params}}_t = f(\Params_{t-1};\bm{\widetilde{\Omega}}) = \widetilde{{\bm A}} {\bm{\Params}}_{t-1}\widetilde{{\bm B}}\approx {\bm \Params}_{t},
\]
where $\widetilde{\bm{\Params}}_t$ is the {\em predicted} matrix of observations at time $t$ according to $f$ when parametrized by coefficient matrices $\bm{\widetilde{\Omega}} = \{\widetilde{{\bm A}},\widetilde{{\bm B}}\}$.
Thus, the key question is how to estimate the best model $f$, namely the best coefficient matrices $\hat{{\bm\Omega}} = \{\hat{{\bm A}},\hat{{\bm B}}\}$.
For starters, we define an instance-level loss function that measures the cost of approximating the true (yet unknown) data generation process with our model $f$ as follows:
\begin{equation}
\begin{split}
\loss(\bm{\widetilde{\Omega}};{\bm \Params}_{t}) & = ||{\bm \Params}_{t} - \widetilde{{\bm \Params}}_{t}||^2_{\text F} \\
& = ||{\bm \Params}_{t} - f(\Params_{t-1};\bm{\widetilde{\Omega}})||^2_{\text F} \\
& = ||{\bm \Params}_{t} - \widetilde{{\bm A}}{\bm \Params}_{t-1}\widetilde{{\bm B}}||^2_{\text F} \\
& = \loss(\widetilde{{\bm A}},\widetilde{{\bm B}}; {\bm \Params}_{t}),
\end{split}
\label{eq:app-inst-loss}
\end{equation}
where $||\cdot||_{\text F}$ indicates the Frobenius norm of a matrix.
More generally, if we have access to $l > 0$ historical matrix observations, we can compute the overall loss function below:
\begin{equation}
\Loss(\widetilde{{\bm A}},\widetilde{{\bm B}}; {\bm \Params}_{t},l) = \sum_{j=0}^{l-1} \loss(\widetilde{{\bm A}},\widetilde{{\bm B}}; {\bm \Params}_{t-j}).
\label{eq:app-loss}
\end{equation}
Notice that $l$ here affects only the size of the training set \textit{not} the order of the autoregressive model. In other words, the forecasting model $f$ will still be MAR($1$) and not MAR($l$), i.e., the matrix of local updates at time $t$ (${\bm \Params}_{t}$) depends \textit{only} on the previously observed matrix at time step $t-1$ (${\bm \Params}_{t-1}$).

Eventually, the best estimates $\hat{\bm A}$ and $\hat{\bm B}$ can be found as the solutions to the following objective:
\begin{equation}
\label{eq:app-mar-opt}
\begin{split}
    \hat{{\bm\Omega}} = \hat{\bm A}, \hat{\bm B} & = \argmin_{\widetilde{{\bm A}}, \widetilde{{\bm B}}}\Big\{\Loss(\widetilde{{\bm A}},\widetilde{{\bm B}}; {\bm \Params}_{t},l) \Big\} \\
    & = \argmin_{\widetilde{{\bm A}}, \widetilde{{\bm B}}}\Big\{\sum_{j=0}^{l-1} ||{\bm \Params}_{t-j} - \widetilde{{\bm A}}{\bm \Params}_{t-j-1}\widetilde{{\bm B}}||^2_{\text F} \Big\}.
\end{split}
\end{equation}
Before solving the optimization task defined in Eq.~(\ref{eq:app-mar-opt}) above, we replace $\widetilde{{\bm A}}$ with ${\bm A}$ and $\widetilde{{\bm B}}$ with ${\bm B}$ to ease the reading. Hence, we observe the following.
A closed-form solution to find $\hat{\bm A}$ can be computed by taking the partial derivative of the loss w.r.t. ${\bm A}$, setting it to $0$, and solving it for ${\bm A}$. In other words, we search for $\hat{\bm A}={\bm A}$, such that:
\begin{equation}
    \label{eq:app-partial-a}
\frac{\partial{\Loss({\bm A},{\bm B}; {\bm \Params}_{t},l)}}{\partial{{\bm A}}} = 0.
\end{equation}

Using Eq.~(\ref{eq:app-loss}) and Eq.~(\ref{eq:app-inst-loss}), the left-hand side of Eq.~(\ref{eq:app-partial-a}) can be rewritten as follows:
\begin{equation}
\label{eq:app-partial}
\small
\begin{split}
\frac{\partial{\Loss({\bm A},{\bm B}; {\bm \Params}_{t},l)}}{\partial{{\bm A}}} & = \frac{\partial{\sum_{j=0}^{l-1} ||{\bm \Params}_{t-j} - {\bm A}{\bm \Params}_{t-j-1}{\bm B}||^2_{\text F}}}{\partial{{\bm A}}}= \\
& = -2 \sum_{j=0}^{l-1}({\bm \Params}_{t-j} - {\bm A}{\bm \Params}_{t-j-1}{\bm B}^T){\bm B}{\bm \Params}^T_{t-1}= \\
& = -2 \Bigg[\Bigg(\sum_{j=0}^{l-1} {\bm \Params}_{t-j}{\bm B}{\bm \Params}^T_{t-j-1}\Bigg) \\
& - {\bm A}\Bigg(\sum_{j=0}^{l-1}{\bm \Params}_{t-j-1}{\bm B}^T{\bm B}{\bm \Params}^T_{t-j-1}\Bigg)\Bigg].
\end{split}
\end{equation}
If we set Eq.~(\ref{eq:app-partial}) to $0$ and solve it for ${\bm A}$, we find:
\begin{equation}
\footnotesize
\begin{split}
    -2 \Bigg[\Bigg(\sum_{j=0}^{l-1} {\bm \Params}_{t-j}{\bm B}{\bm \Params}^T_{t-j-1}\Bigg) - {\bm A}\Bigg(\sum_{j=0} ^{l-1}{\bm \Params}_{t-j-1}{\bm B}^T{\bm B}{\bm \Params}^T_{t-j-1}\Bigg)\Bigg] = 0 \nonumber
\end{split}
\end{equation}
 \[ \iff \]
\begin{equation}
\footnotesize
\begin{split}
    \Bigg(\sum_{j=0}^{l-1} {\bm \Params}_{t-j}{\bm B}{\bm \Params}^T_{t-j-1}\Bigg) - {\bm A}\Bigg(\sum_{j=0}^{l-1}{\bm \Params}_{t-j-1}{\bm B}^T{\bm B}{\bm \Params}^T_{t-j-1}\Bigg) = 0. \nonumber
\end{split}
\end{equation}

Hence:

\begin{equation}
\footnotesize
\begin{split}
\Bigg(\sum_{j=0}^{l-1} {\bm \Params}_{t-j}{\bm B}{\bm \Params}^T_{t-j-1}\Bigg)  
- {\bm A}\Bigg(\sum_{j=0}^{l-1}{\bm \Params}_{t-j-1}{\bm B}^T{\bm B}{\bm \Params}^T_{t-j-1}\Bigg) = 0
\end{split}
\end{equation}

\begin{equation}
\footnotesize
\begin{split}
\Bigg(\sum_{j=0}^{l-1} {\bm \Params}_{t-j}{\bm B}{\bm \Params}^T_{t-j-1}\Bigg) = 
{\bm A}\Bigg(\sum_{j=0}^{l-1}{\bm \Params}_{t-j-1}{\bm B}^T{\bm B}{\bm \Params}^T_{t-j-1}\Bigg) \label{eq:app-A-imp}
\end{split}
\end{equation}

\begin{equation}
\footnotesize
\begin{split}
{\bm A} = \Bigg(\sum_{j=0}^{l-1} {\bm \Params}_{t-j}{\bm B}{\bm \Params}^T_{t-j-1}\Bigg) 
\Bigg(\sum_{j=0}^{l-1}{\bm \Params}_{t-j-1}{\bm B}^T{\bm B}{\bm \Params}^T_{t-j-1}\Bigg)^{-1}.
\label{eq:app-A}
\end{split}
\end{equation}

Notice that Eq.~(\ref{eq:app-A}) is obtained by multiplying both sides of Eq.~(\ref{eq:app-A-imp}) by $\Big(\sum_{j=0}^{l-1}{\bm \Params}_{t-j-1}{\bm B}^T{\bm B}{\bm \Params}^T_{t-j-1}\Big)^{-1}$.

If we apply the same reasoning, we can also find a closed-form solution to compute $\hat{\bm B}$. That is, we take the partial derivative of the loss w.r.t. ${\bm B}$, set it to $0$, and solve it for ${\bm B}$:
\begin{equation}
\label{eq:app-partial-b}
\frac{\partial{\Loss({\bm A},{\bm B}; {\bm \Params}_{t},l)}}{\partial{{\bm B}}} = 0.
\end{equation}
Eventually, we obtain the following:
\begin{equation}
\small
\begin{split}
\label{eq:app-B}
{\bm B} = \Bigg(\sum_{j=0}^{l-1} {\bm \Params}^T_{t-j}{\bm A}{\bm \Params}_{t-j-1}\Bigg) 
\Bigg(\sum_{j=0}^{l-1}{\bm \Params}^T_{t-j-1}{\bm A}^T{\bm A}{\bm \Params}_{t-j-1}\Bigg)^{-1}.
\end{split}
\end{equation}
We now have two closed-form solutions; one for ${\bm A}$ (see Eq.~(\ref{eq:app-A})) and one for ${\bm B}$ (see Eq.~(\ref{eq:app-B})).
However, the solution to ${\bm A}$ involves ${\bm B}$, and the solution to ${\bm B}$ involves ${\bm A}$. In other words, we must know ${\bm B}$ to compute ${\bm A}$ and vice versa.

We can use the standard Alternating Least Squares (ALS) algorithm (\cite{koren2009ieeecomp}) to solve such a problem. 
The fundamental idea of ALS is to iteratively update the least squares closed-form solution of each variable alternately, keeping the other fixed. At the generic $i$-th iteration, we compute:
\begin{equation}
\scriptsize
\begin{split}
    {\bm A}^{(i+1)} = \Bigg(\sum_{j=0}^{l-1} {\bm \Params}_{t-j}{\bm B}^{(i)}{\bm \Params}^T_{t-j-1}\Bigg) \Bigg(\sum_{j=0}^{l-1}{\bm \Params}_{t-j-1}({\bm B}^{(i)})^T{\bm B}^{(i)}{\bm \Params}^T_{t-j-1}\Bigg)^{-1};\nonumber
\end{split}
\end{equation}
\begin{equation}
\scriptsize
\begin{split}
    {\bm B}^{(i+1)} = \Bigg(\sum_{j=0}^{l-1} {\bm \Params}^T_{t-j}{\bm A}^{(i+1)}{\bm \Params}_{t-j-1}\Bigg) \Bigg(\sum_{j=0}^{l-1}{\bm \Params}^T_{t-j-1}({\bm A}^{(i+1)})^T{\bm A}^{(i+1)}{\bm \Params}_{t-j-1}\Bigg)^{-1};\nonumber
\end{split}
\end{equation}
ALS repeats the two steps above until some convergence criterion is met, e.g., after a specific number of iterations $N$ or when the distance between the values of the variables computed in two consecutive iterations is smaller than a given positive threshold, i.e., $d({\bm A}^{(i+1)}-{\bm A}^{(i)}) < \varepsilon$ and $d({\bm B}^{(i+1)}-{\bm B}^{(i)}) < \varepsilon$, where $d(\cdot)$ is any suitable matrix distance function and $\varepsilon \in \R_{>0}$.

Eventually, if ${\bm A}^{(\infty)}$ and ${\bm B}^{(\infty)}$ are the parameters of the MAR model upon convergence, we set $\hat{\bm A} = {\bm A}^{(\infty)}$ and $\hat{\bm B} = {\bm B}^{(\infty)}$ as the best coefficient matrices.

\section{Datasets, Tasks, and FL Models}
\label{app:setup}

In Table~\ref{tab:setup}, we report the full details of our experimental setup concerning the datasets used, their associated tasks, and the models (along with their hyperparameters) trained on the simulated FL environment, i.e., Flower\footnote{\url{https://flower.dev/}} (see Appendix~\ref{app:fl-env}). Table~\ref{tab:hyp} shows the description of the hyperparameters.

\begin{table*}[ht]
\centering
\caption{Experimental setup: datasets, tasks, and FL models considered.
}
\label{tab:setup}
\scalebox{1}{
\begin{tabular}{cccccc}
\toprule
\textbf{Dataset}        & \thead{\textbf{N. of Instances}\\\textbf{(training/test)}} & \textbf{N. of Features} & \textbf{Task} & \textbf{FL Model} & \textbf{Hyperparameters}\\
\midrule
\thead{\textit{MNIST}~\cite{mnist-ds}} & $60,000$/$10,000$ & \thead{$28$x$28$\\(numerical)} & \thead{multiclass\\classification} & MLP & \thead{\{{\tt batch}=$32$; {\tt layers}=$2$;\\{\tt opt}=Adam; $\eta$=$10^{-3}$\}}\\
\midrule
\thead{\textit{Fashion-MNIST}~\cite{fashionmnist-ds}} & $60,000$/$10,000$ & \thead{$28$x$28$\\(numerical)} & \thead{multiclass\\classification} & MLP & \thead{\{{\tt batch}=$32$; {\tt layers}=$4$;\\{\tt opt}=Adam; $\eta$=$10^{-3}$\}}\\
\midrule
\thead{\textit{CIFAR-10}~\cite{cifar10-100-ds}} & $50,000$/$10,000$ & \thead{$32$x$32$x$3$\\(numerical)} & \thead{multiclass\\classification} & CNN & \thead{\{{\tt batch}=$32$; {\tt layers}=$6$;\\{\tt opt}=SGD; $\eta$=$10^{-2}$; $\mu$=$0.9$\}}\\
\midrule
\thead{\textit{CIFAR-100}~\cite{cifar10-100-ds}} & $50,000$/$10,000$ & \thead{$32$x$32$x$3$\\(numerical)} & \thead{multiclass\\classification} &  CNN (MobileNet) & \thead{\{{\tt batch}=$32$; {\tt layers}=$28$;\\{\tt opt}=SGD; $\eta$=$10^{-2}$; $\mu$=$0.9$\}}\\
\bottomrule
\end{tabular}
}
\end{table*}

\begin{table}[ht!]
\centering
\caption{Description of hyperparameters.
}
\label{tab:hyp}
\scalebox{1}{
\begin{tabular}{cc}
    \toprule
    \textbf{Hyperparameter}        & \textbf{Description} \\ 
    \midrule
    $\eta$ & learning rate\\
    \midrule
    $\mu$ & momentum\\
    \midrule
    {\tt opt} & optimizer: stochastic gradient descent (SGD); Adam\\
    \midrule
    {\tt batch} & batch size\\
    \midrule
    {\tt layers} & number of neural network layers\\
    \bottomrule
    \end{tabular}
}
\end{table}

The MLP for the \textit{MNIST} dataset is a $2$-layer fully connected feed-forward neural network, whereas the MLP for the \textit{Fashion-MNIST} dataset is a $4$-layer fully connected feed-forward neural network. Both MLPs are trained by minimizing multiclass cross-entropy loss using Adam optimizer with batch size equal to $32$~\cite{li2022blades}.
The CNN used for \textit{CIFAR-10} is a $6$-layer convolutional neural network, while the CNN used for \textit{CIFAR-100} is the well-known MobileNet architecture~\cite{howard2017mobilenets}. Both CNNs are trained by minimizing multiclass cross-entropy loss via stochastic gradient descent (SGD) with batch size equal to $32$~\cite{pytorch-cifar-10}.

At each FL round, every client performs one training epoch of Adam/SGD, which corresponds to the number of iterations needed to ``see'' all the training instances once, when divided into batches of size $32$.
In any case, the updated local model is sent to the central server for aggregation.

We run our experiments on a machine equipped with an AMD Ryzen 9, 64 GB RAM, and an NVIDIA 4090 GPU with 24 GB VRAM.

\section{FL Simulation Environment}
\label{app:fl-env}
To simulate a realistic FL environment, we integrate FLANDERS into Flower. Moreover, we implement in Flower every other defense baseline considered in this work that the framework does not natively provide.
Other valid FL frameworks are available (e.g., TensorFlow Federated\footnote{\scriptsize{\url{https://www.tensorflow.org/federated}}} and PySyft\footnote{\scriptsize{\url{https://github.com/OpenMined/PySyft}}}), but Flower turned out the most flexible.

We want to remark that within Flower, only the \textit{number} of malicious clients in each FL round remains constant, while the framework manages their selection. This implies that a single client can alternate between legitimate and malicious behavior over successive FL rounds.


We report the main properties of our FL environment simulated on Flower in Table~\ref{tab:fl-env}.

\section{Defense Settings}
\label{app:defenses}
Below, we describe the critical parameters for
each non-trivial baseline considered.

\noindent{{\bf {\em Trimmed Mean}}{\bf.}} 
The key parameter of this defense strategy is $\beta$, which is used to cut the parameters on the edges. In this work, we set $\beta = 0.2$.

\noindent{{\bf {\em FedMedian}}{\bf.}}
This method has no parameters to set.

\noindent{{\bf {\em Multi-Krum}}{\bf.}} 
We set the number of local models to keep for the aggregation (FedAvg) after the Krum filtering as $k=(b-m)$

\noindent{{\bf {\em Bulyan}}{\bf.}}
The two crucial parameters of this hybrid robust aggregation rule are $\alpha$ and $\beta$. The former determines the number of times Krum is applied to generate $\alpha$ local models; the latter is used to determine the number of parameters to select closer to the median.
In this work, we set $\alpha = m - 2 \cdot b, \beta = \alpha - 2 \cdot b$, where $b=\{0,20\}$.

\noindent{{\bf {\em DnC}}{\bf.}}
This is an iterative algorithm that has three parameters: we set $niters = 5$ as suggested by the authors, and the filtering fraction $c = 1$ to keep exactly one model (i.e., the one with the best anomaly score). Furthermore, we set $\widetilde{d} = 500$ on \textit{CIFAR-100} to sample $500$ local model weights.

\noindent{{\bf {\em FLDetector}}{\bf.}}
We set the window size $N=20$, and we let FLDetector know how many local models to keep, i.e., $k=(m-b)$, as we did for the other baselines.



\noindent{{\bf {\em FLANDERS}}{\bf.}}
For a fair comparison with other baselines, we set the sliding window size $l=2$, and the number of clients to keep at every round $k=(m-b)$, where $m$ is the number of clients selected at each round and $b$ the total number of malicious clients in the FL system. Furthermore, we use a sampling value $\widetilde{d}\leq d$ that indicates how many parameters we store in the history, the number $N$ of ALS iterations, and $\alpha$ and $\beta$ which are regularization factors. Random sampling is used to select the subset of parameters considered. This is a common strategy proposed in the literature~\cite{shejwalkar2021ndss} to lower the model size to train and save the server's memory. On the other hand, the regularization factors are needed when the model parameters, the number of clients selected, or $\widetilde{d}$ cause numerical problems in the ALS algorithm, whose number of iterations is set to $N=100$. We always set $\alpha=1$ and $\beta=1$, meaning that there is no regularization since, in our experience, it reduces the capability of predicting the right model. For this reason, we set $\widetilde{d}=500$.
Finally, we set the distance function $\delta$ used to measure the difference between the observed vector of weights sent to the server and the predicted vector of weights output by MAR to squared $L^2$-norm. 
In future work, we plan to investigate other distance measures, such as cosine or mutual information distance.

Table~\ref{tab:defenses} summarizes the values of the key parameters discussed above and those characterizing our method FLANDERS.
\begin{table*}[h]
\centering
\caption{Key parameter settings for each defense strategy considered.
}
\label{tab:defenses}
\vspace{2mm}
\scalebox{0.95}{
    \begin{tabular}{cc}
    \toprule
    \textbf{Defense}        & \textbf{Parameters} \\ 
    \midrule
    {\bf {\em FedAvg}} & N/A\\ 
    \midrule
    {\bf {\em Trimmed Mean}} & $\beta=0.2$\\ 
    \midrule
    {\bf {\em FedMedian}} & N/A\\ 
    \midrule
    {\bf {\em Multi-Krum}} & $b=\{0,20\}; k=(m-b)$\\ 
    \midrule
    {\bf {\em Bulyan}} & $b=\{0,20\}; \alpha = m - 2 \cdot b; \beta = \alpha - 2 \cdot b$\\ 
    \midrule
    {\bf {\em DnC}} & $\widetilde{d}=500; niters=5; c=1$ \\ 
    \midrule
    {\bf {\em FLDetector}} & $N=20; k=(m-b)$ \\
    \midrule
    {\bf {\em FLANDERS}} & $l=2$; $k=(m-b)$; $\widetilde{d}=\{0, 500\}$; $N=100$; $\alpha=1$; $\beta=1$; $\delta = \text{squared } L^2$-norm  \\
    \bottomrule
    \end{tabular}
}
\end{table*}

\section{Additional Results}
\label{app:further-results}

\subsection{Impact on Attack-Free scenarios}
First of all, in Table~\ref{tab:increment-0}, we assess the impact of FLANDERS in an attack-free scenario (i.e., when $r=0$). In this setting, no clear winning strategy emerges. Sometimes, FLANDERS has a detrimental effect on the global model's accuracy with a standard aggregation mechanism (e.g., see FedAvg with the \textit{MNIST} dataset). In other instances, however, FLANDERS improves the global model's accuracy when paired with robust aggregation schemes (e.g., see Bulyan with the \textit{MNIST} dataset).

\begin{table}[htb!]
\centering
\caption{Accuracy of the global model using all the baseline aggregation methods without and with FLANDERS ($r=0$). The best results are typed in boldface.}
\label{tab:increment-0}
\scalebox{1}{
\begin{tabular}{ccccc}
\toprule
& \multicolumn{1}{c}{\textit{MNIST}} & \multicolumn{1}{c}{\textit{Fashion-MNIST}} & \multicolumn{1}{c}{\textit{CIFAR-10}} & \multicolumn{1}{c}{\textit{CIFAR-100}} \\
\midrule
FedAvg                  & ${\bf 0.86}$ & $0.63$ & $0.35$ & ${\bf 0.06}$ \\
+ FLANDERS              & $0.83$ & ${\bf 0.68}$ & ${\bf 0.36}$ & $0.05$ \\
\midrule
FedMedian               & ${\bf 0.84}$ & ${\bf 0.71}$ & $0.31$ & ${\bf 0.10}$ \\
+ FLANDERS              & $0.78$ & $0.68$ & ${\bf 0.33}$ & ${\bf 0.10}$ \\
\midrule
TrimmedMean             & ${\bf 0.82}$ & $0.69$ & ${\bf 0.33}$ & ${\bf 0.12}$ \\
+ FLANDERS              & $0.80$ & ${\bf 0.70}$ & $0.32$ & $0.11$ \\
\midrule
MultiKrum               & $0.72$ & ${\bf 0.65}$ & $0.34$ & $0.05$ \\
+ FLANDERS              & ${\bf 0.80}$ & $0.64$ & ${\bf 0.44}$ & ${\bf 0.08}$ \\
\midrule
Bulyan                  & $0.85$ & $0.65$ & ${\bf 0.43}$ & $0.05$ \\
+ FLANDERS              & ${\bf 0.90}$ & ${\bf 0.70}$ & $0.42$ & ${\bf 0.06}$ \\
\midrule
DnC                     & $0.81$ & $0.62$ & ${\bf 0.44}$ & $0.06$ \\
+ FLANDERS              & ${\bf 0.86}$ & ${\bf 0.64}$ & ${\bf 0.44}$ & ${\bf 0.07}$ \\
\bottomrule
\end{tabular}
}
\end{table}

\subsection{Random Client Selection}
\label{app:client-selection}
So far, we have assumed that the FL server selects \textit{all} available clients at each round, i.e., $|\mathcal{C}^{(t)}| = m = K~\forall t\in \{1,2, \ldots, T\}$.
In this section, we investigate the scenario where the number of clients selected at each round remains fixed ($|\mathcal{C}^{(t)}| = m$), but now the FL server chooses a \textit{random} subset of the available clients, i.e., $1 \leq m < K$.
\\
We experiment with $K=10$ clients, of which $b=2$ are malicious (running the AGR-MM attack). Each client has a non-iid sample of the \textit{MNIST} dataset, where we fix $k=max(1,K*c - b)$. At each FL round, $m=K*c$ clients are randomly chosen ($c \in \{0.2,0.5,1.0\}$). 
\begin{table}[htb!]
    \centering
    \caption{Accuracy of the global model on the \textit{MNIST} dataset using FLANDERS + FedAvg, with $b=2$ attackers out of $K=10$ clients, under AGR-MM attack, and with variable clients selected for training each round ($c$).}
    \label{tab:rnd-client-sel}
    \begin{tabular}{cccc}
    \toprule
    $c$ & ${\bf 0.2}$ & ${\bf 0.5}$ & ${\bf 1.0}$ \\
    \midrule
    Accuracy & $0.75$ & $0.83$ & $0.92$ \\
    \bottomrule
    \end{tabular}
    \label{tab:random-client-selection}
\end{table}
\\
The results in Table~\ref{tab:random-client-selection} show that the accuracy of the global model improves as the ratio of sampled clients increases, while FLANDERS remains robust to AGR-MM attacks.

\subsection{Malicious Detection Accuracy}
\label{app:flanders-accuracy}
We evaluate the capability of FLANDERS to detect malicious clients accurately. 
Specifically, let $\mathcal{C}^{(t)}_{\text{mal}}\subseteq \mathcal{C}$ be the set of malicious clients selected at round $t$ by the FL server. 
Furthermore, let $\hat{\mathcal{C}}^{(t)}_{\text{mal}}\subseteq \mathcal{C}$ be the set of malicious clients identified by FLANDERS at round $t$. Thus, we measure \textit{Precision} ($P$) and \textit{Recall} ($R$) as usual, i.e., $P=\frac{tp}{tp + fp}$ and $R=\frac{tp}{tp + fn}$, where $tp,fp,fn$ stand for \textit{true positives}, \textit{false positives}, and \textit{false negatives}. 
Specifically, $tp = \sum_{t=1}^T |\mathcal{C}^{(t)}_{\text{mal}} \cap \hat{\mathcal{C}}^{(t)}_{\text{mal}}|$, $fp = \sum_{t=1}^T |\hat{\mathcal{C}}^{(t)}_{\text{mal}} \setminus \mathcal{C}^{(t)}_{\text{mal}}|$, and $fn = \sum_{t=1}^T |\mathcal{C}^{(t)}_{\text{mal}} \setminus \hat{\mathcal{C}}^{(t)}_{\text{mal}}|$.

Table~\ref{tab:pr-60} illustrates the values for the precision ($P$) and recall ($R$) values of FLANDERS under extreme attack scenarios, integrating the findings already reported in Tables~\ref{tab:pr-20}, and \ref{tab:pr-80} in the main body.

\begin{table*}[htb!]
\centering
\caption{\textit{Precision} ($P$) and \textit{Recall} ($R$) of FLDetector and FLANDERS in detecting malicious clients across various datasets and attacks ($r=0.6$; $T=50$ rounds).}
\label{tab:pr-60}
\scalebox{1}{
\begin{tabular}{@{}cccccccccccccccccc@{}}\toprule
 \multicolumn{1}{c}{} & \multicolumn{8}{c}{\textbf{FLDetector}} & \phantom{a} & \multicolumn{8}{c}{\textbf{FLANDERS}} \\
 \cmidrule(lr){2-9} \cmidrule(lr){11-18}
 & \multicolumn{2}{c}{GAUSS} & \multicolumn{2}{c}{LIE} & \multicolumn{2}{c}{OPT} & \multicolumn{2}{c}{AGR-MM} && \multicolumn{2}{c}{GAUSS} & \multicolumn{2}{c}{LIE} & \multicolumn{2}{c}{OPT} & \multicolumn{2}{c}{AGR-MM} \\
 \cmidrule(lr){2-3} \cmidrule(lr){4-5} \cmidrule(lr){6-7} \cmidrule(lr){8-9} \cmidrule(lr){11-12} \cmidrule(lr){13-14} \cmidrule(lr){15-16} \cmidrule(lr){17-18}
                            & $P$   & $R$   & $P$   & $R$   & $P$    & $R$    & $P$   & $R$     && $P$   & $R$   & $P$   & $R$    & $P$  & $R$   & $P$  & $R$ \\
\midrule
\textit{MNIST}         & $0.60$ & $0.60$ & $0.60$ & $0.60$ & $0.62$ & $0.62$ & $0.60$ & $0.60$   && ${\bf 1.0}$ & ${\bf 1.0}$ & ${\bf 1.0}$ & ${\bf 1.0}$ & ${\bf 1.0}$ & ${\bf 1.0}$ & ${\bf 1.0}$ & ${\bf 1.0}$ \\
\textit{Fashion-MNIST} & $0.60$ & $0.60$ & $0.59$ & $0.59$ & $0.60$ & $0.60$ & $0.61$ & $0.61$   && ${\bf 1.0}$ & ${\bf 1.0}$ & ${\bf 1.0}$ & ${\bf 1.0}$ & ${\bf 1.0}$ & ${\bf 1.0}$ & ${\bf 1.0}$ & ${\bf 1.0}$ \\
\textit{CIFAR-10}      & $0.60$ & $0.60$ & $0.61$ & $0.61$ & $0.60$ & $0.60$ & $0.60$ & $0.60$   && ${\bf 1.0}$ & ${\bf 1.0}$ & ${\bf 1.0}$ & ${\bf 1.0}$ & ${\bf 1.0}$ & ${\bf 1.0}$ & ${\bf 1.0}$ & ${\bf 1.0}$ \\
\textit{CIFAR-100}     & $0.60$ & $0.60$ & $0.60$ & $0.60$ & $0.60$ & $0.60$ & $0.61$ & $0.61$   && ${\bf 1.0}$ & ${\bf 1.0}$ & ${\bf 1.0}$ & ${\bf 1.0}$ & ${\bf 1.0}$  & ${\bf 1.0}$  & ${\bf 1.0}$ & ${\bf 1.0}$ \\
\bottomrule
\end{tabular}
}
\end{table*}

\subsection{Aggregation Robustness Lift}
\label{app:agg-lift}

In this section, we report the full results on the improved robustness of the global model against malicious attacks when FLANDERS is paired with existing aggregation strategies.

Table~\ref{tab:increment-fld-60} illustrates the comparison between FLANDERS and its main competitor, FLDetector, when both are paired with standard FedAvg, under a severe ($r=0.6$) attack scenario. These results complete those shown in Table~\ref{tab:increment-fld}, and \ref{tab:increment-fld-60} of the main submission.


\begin{table*}[htb!]
\centering
\caption{Accuracy of the global model using FedAvg with FLDetector and FLANDERS ($r=0.6$).}
\label{tab:increment-fld-60}
\scalebox{1}{
\begin{tabular}{ccccccccccccccc}
\toprule
& \multicolumn{4}{c}{\textbf{FedAvg}} & \phantom{a} & \multicolumn{4}{c}{\textbf{FLDetector + FedAvg}} & \phantom{a} & \multicolumn{4}{c}{\textbf{FLANDERS + FedAvg}} \\
\cmidrule(lr){2-5} \cmidrule(lr){7-10} \cmidrule(lr){12-15}
                    &  GAUSS    &  LIE     &  OPT     &  AGR-MM  &&  GAUSS   &  LIE     &  OPT     &  AGR-MM  &&  GAUSS    &  LIE     &  OPT     &  AGR-MM      \\ 
\midrule
\textit{MNIST}          & $0.18$ & $0.12$ & $0.16$ & $0.15$ && $0.19$ & $0.12$ & $0.17$ & $0.12$ && ${\bf 0.78}$ & ${\bf 0.82}$ & ${\bf 0.79}$ & ${\bf 0.85}$ \\
\textit{Fashion-MNIST}  & $0.28$ & $0.10$ & $0.18$ & $0.10$ && $0.19$ & $0.10$ & $0.11$ & $0.10$ && ${\bf 0.69}$ & ${\bf 0.65}$ & ${\bf 0.65}$ & ${\bf 0.62}$ \\
\textit{CIFAR-10}       & $0.10$ & $0.10$ & $0.10$ & $0.10$ && $0.10$ & $0.10$ & $0.10$ & $0.10$ && ${\bf 0.36}$ & ${\bf 0.36}$ & ${\bf 0.34}$ & ${\bf 0.35}$ \\
\textit{CIFAR-100}      & $0.01$ & $0.01$ & $0.01$ & $0.01$ && $0.01$ & $0.01$ & $0.01$ & $0.01$ && ${\bf 0.08}$ & ${\bf 0.09}$ & ${\bf 0.08}$ & ${\bf 0.09}$ \\
\bottomrule
\end{tabular}
}
\end{table*}

Finally, Tables~\ref{tab:increment-20}, and \ref{tab:increment-60} illustrate the impact of FLANDERS on the global model's accuracy under light ($r=0.2$) and severe ($r=0.6$) attack settings.

\begin{table*}[htb!]
\centering
\caption{Accuracy of the global model using all the baseline aggregation methods without and with FLANDERS ($r=0.2$). The best results are typed in boldface.}
\label{tab:increment-20}
\scalebox{0.75}{
\begin{tabular}{cccccccccccccccccccc}
\toprule
& \multicolumn{4}{c}{\textit{MNIST}} & \phantom{a} & \multicolumn{4}{c}{\textit{Fashion-MNIST}} & \phantom{a} & \multicolumn{4}{c}{\textit{CIFAR-10}} & \phantom{a} & \multicolumn{4}{c}{\textit{CIFAR-100}} \\
\cmidrule(lr){2-5} \cmidrule(lr){7-10} \cmidrule(lr){12-15} \cmidrule(lr){17-20}
              &  GAUSS        &  LIE          &  OPT          &  AGR-MM       &&  GAUSS        &  LIE          &  OPT          &  AGR-MM       &&  GAUSS         &  LIE          &  OPT          &  AGR-MM        &&  GAUSS         &  LIE          &  OPT          &  AGR-MM      \\ 
\midrule
FedAvg        &  $0.18$       &  $0.12$       &  ${\bf 0.63}$ &  $0.34$       &&  $0.25$       & $0.10$        &  $0.56$       &  $0.16$       &&  $0.10$       &  $0.10$       &  $0.23$       &  $0.10$       &&  $0.01$       &  $0.01$       &  $0.01$       &  $0.02$  \\
+ FLANDERS    &  ${\bf 0.86}$ &  ${\bf 0.83}$ &  $0.62$       &  ${\bf 0.85}$ &&  ${\bf 0.69}$ & ${\bf 0.64}$  &  ${\bf 0.58}$ &  ${\bf 0.63}$ &&  ${\bf 0.38}$ &  ${\bf 0.37}$ &  ${\bf 0.28}$ &  ${\bf 0.36}$ &&  ${\bf 0.07}$ &  ${\bf 0.05}$ &  ${\bf 0.05}$ &  ${\bf 0.06}$  \\
\midrule
FedMedian     &  ${\bf 0.81}$ &  $0.57$       &  ${\bf 0.61}$ &  $0.63$       &&  $0.70$       &  $0.63$       &  ${\bf 0.68}$ &  $0.62$       &&  ${\bf 0.34}$ &  $0.24$       &  ${\bf 0.24}$ &  $0.21$       &&  $0.01$       &  $0.04$       &  $0.01$       &  $0.03$  \\
+ FLANDERS    &  ${\bf 0.81}$ &  ${\bf 0.86}$ &  $0.58$       &  ${\bf 0.85}$ &&  ${\bf 0.71}$ &  ${\bf 0.73}$ &  $0.64$       &  ${\bf 0.72}$ &&  ${\bf 0.34}$ &  ${\bf 0.28}$ &  $0.17$       &  ${\bf 0.32}$ &&  ${\bf 0.10}$       &  ${\bf 0.10}$ &  ${\bf 0.10}$ &  ${\bf 0.11}$  \\
\midrule
TrimmedMean   &  $0.78$       &  $0.57$       &  ${\bf 0.76}$ &  $0.48$       &&  $0.65$       &  $0.58$       &  $0.61$       &  $0.55$       &&  ${\bf 0.35}$ &  $0.24$       &  ${\bf 0.23}$ &  $0.23$       &&  $0.01$       &  $0.03$       &  $0.01$       &  $0.03$  \\
+ FLANDERS    &  ${\bf 0.83}$ &  ${\bf 0.82}$ &  $0.75$       &  ${\bf 0.77}$ &&  ${\bf 0.69}$ &  ${\bf 0.69}$ &  ${\bf 0.63}$ &  ${\bf 0.70}$ &&  ${\bf 0.35}$ &  ${\bf 0.33}$ &  $0.22$       &  ${\bf 0.35}$ &&  ${\bf 0.11}$       &  ${\bf 0.10}$ &  ${\bf 0.10}$ &  ${\bf 0.10}$  \\
\midrule
Multi-Krum    &  $0.73$       &  ${\bf 0.86}$ &  $0.83$       &  ${\bf 0.80}$ &&  ${\bf 0.66}$ &  ${\bf 0.67}$ &  $0.65$       &  $0.64$       &&  $0.36$       &  $0.35$       &  $0.27$       &  $0.34$       &&  $0.07$       &  $0.06$       &  $0.07$       &  $0.07$  \\
+ FLANDERS    &  ${\bf 0.87}$ &  $0.85$       &  ${\bf 0.85}$ &  $0.77$       &&  $0.65$       &  $0.64$       &  ${\bf 0.69}$ &  ${\bf 0.66}$ &&  ${\bf 0.42}$ &  ${\bf 0.44}$ &  ${\bf 0.35}$ &  ${\bf 0.42}$ &&  ${\bf 0.10}$ &  ${\bf 0.09}$ &  ${\bf 0.10}$ &  ${\bf 0.09}$  \\
\midrule
Bulyan        &  $0.81$       &  ${\bf 0.88}$ &  ${\bf 0.85}$ &  $0.84$       &&  ${\bf 0.67}$ &  ${\bf 0.72}$ &  ${\bf 0.70}$ &  ${\bf 0.76}$ &&  $0.40$       &  $0.39$       &  $0.34$       &  $0.39$       &&  ${\bf 0.10}$ &  ${\bf 0.10}$ &  ${\bf 0.10}$ &  ${\bf 0.08}$  \\
+ FLANDERS    &  ${\bf 0.83}$ &  ${\bf 0.88}$ &  $0.84$       &  ${\bf 0.87}$ &&  $0.62$       &  $0.66$       &  $0.66$       &  $0.68$       &&  ${\bf 0.43}$ &  ${\bf 0.43}$ &  ${\bf 0.36}$ &  ${\bf 0.42}$ &&  $0.07$       &  $0.06$       &  $0.06$       &  ${\bf 0.08}$  \\
\midrule
DnC           &  $0.20$       &  $0.14$       &  ${\bf 0.69}$ &  $0.30$       &&  $0.25$       &  $0.18$       &  $0.59$       &  $0.17$       &&  $0.11$       &  $0.10$       &  $0.27$       &  $0.15$       &&  $0.01$       &  $0.01$       &  $0.01$       &  $0.02$  \\
+ FLANDERS    &  ${\bf 0.88}$ &  ${\bf 0.87}$ &  $0.65$       &  ${\bf 0.88}$ &&  ${\bf 0.61}$ &  ${\bf 0.64}$ &  ${\bf 0.62}$ &  ${\bf 0.66}$ &&  ${\bf 0.43}$ &  ${\bf 0.44}$ &  ${\bf 0.34}$ &  ${\bf 0.43}$ &&  ${\bf 0.07}$ &  ${\bf 0.07}$ &  ${\bf 0.07}$ &  ${\bf 0.08}$  \\
\bottomrule
\end{tabular}
}
\end{table*}

\begin{table*}[htb!]
\centering
\caption{Accuracy of the global model using all the baseline aggregation methods without and with FLANDERS ($r=0.6$). The best results are typed in boldface.}
\label{tab:increment-60}
\scalebox{0.75}{
\begin{tabular}{cccccccccccccccccccc}
\toprule
& \multicolumn{4}{c}{\textit{MNIST}} & \phantom{a} & \multicolumn{4}{c}{\textit{Fashion-MNIST}} & \phantom{a} & \multicolumn{4}{c}{\textit{CIFAR-10}} & \phantom{a} & \multicolumn{4}{c}{\textit{CIFAR-100}} \\
\cmidrule(lr){2-5} \cmidrule(lr){7-10} \cmidrule(lr){12-15} \cmidrule(lr){17-20}
              &  GAUSS        &  LIE          &  OPT          &  AGR-MM       &&  GAUSS        &  LIE          &  OPT          &  AGR-MM       &&  GAUSS         &  LIE          &  OPT          &  AGR-MM        &&  GAUSS         &  LIE          &  OPT          &  AGR-MM      \\ 
\midrule
FedAvg      & $0.18$        & $0.12$        & $0.16$        & $0.15$        && $0.28$       & $0.10$        & $0.18$        & $0.10$        && $0.10$       & $0.10$        & $0.10$        && $0.10$       & $0.01$       & $0.01$    & $0.01$      & $0.01$ \\
+ FLANDERS  & ${\bf 0.78}$  & ${\bf 0.82}$  & ${\bf 0.79}$  & ${\bf 0.85}$  && ${\bf 0.69}$ & ${\bf 0.65}$  & ${\bf 0.65}$  & ${\bf 0.62}$  && ${\bf 0.36}$ & ${\bf 0.36}$  & ${\bf 0.34}$  && ${\bf 0.35}$ & ${\bf 0.08}$ & ${\bf 0.09}$ & ${\bf 0.08}$ & ${\bf 0.09}$ \\ 
\midrule
FedMedian   & ${\bf 0.84}$  & $0.18$        & $0.12$        & $0.16$        && $0.70$       & $0.10$        & $0.13$        & $0.10$        && ${\bf 0.34}$ & $0.10$        & $0.11$        && $0.10$   & $0.01$         & $0.01$        & $0.01$    & $0.01$ \\
+ FLANDERS  & $0.73$        & ${\bf 0.81}$  & ${\bf 0.83}$  & $0{\bf .82}$  && ${\bf 0.72}$ & ${\bf 0.73}$  & ${\bf 0.71}$  & ${\bf 0.72}$  && $0.31$       & ${\bf 0.31}$  & ${\bf 0.30}$  && ${\bf 0.33}$ & ${\bf 0.11}$ & ${\bf 0.10}$ & ${\bf 0.11}$ & ${\bf 0.10}$ \\
\midrule
TrimmedMean & $0.22$        & $0.12$        & $0.21$        & $0.13$        && $0.33$       & $0.10$        & $0.18$        & $0.10$        && $0.12$       & $0.10$        & $0.10$        && $0.10$       & $0.01$       & $0.01$    & $0.01$      & $0.01$ \\
+ FLANDERS  & ${\bf 0.76}$  & ${\bf 0.83}$  & ${\bf 0.83}$  & ${\bf 0.78}$  && ${\bf 0.72}$ & ${\bf 0.70}$  & ${\bf 0.71}$  & ${\bf 0.68}$  && ${\bf 0.35}$ & ${\bf 0.28}$  & ${\bf 0.33}$  && ${\bf 0.29}$ & ${\bf 0.10}$ & ${\bf 0.11}$ & ${\bf 0.11}$ & ${\bf 0.12}$ \\
\midrule
Multi-Krum  & $0.85$        & $0.24$        & $0.27$        & $0.12$        && $0.65$       & $0.10$        & $0.11$        & $0.10$        && $0.10$       & $0.35$        & $0.10$        && $0.13$       & $0.09$       & $0.01$    & $0.01$      & $0.01$ \\
+ FLANDERS  & ${\bf 0.89}$  & ${\bf 0.84}$  & ${\bf 0.84}$  & ${\bf 0.88}$  && ${\bf 0.71}$ & ${\bf 0.70}$  & ${\bf 0.68}$  & ${\bf 0.67}$  && ${\bf 0.41}$ & ${\bf 0.40}$  & ${\bf 0.40}$  && ${\bf 0.40}$ & ${\bf 0.12}$ & ${\bf 0.12}$ & ${\bf 0.11}$ & ${\bf 0.10}$ \\
\midrule
Bulyan      & N/A           & N/A           & N/A           & N/A           && N/A          & N/A           & N/A           & N/A           && N/A          & N/A           & N/A           && N/A & N/A & N/A & N/A & N/A \\
+ FLANDERS  & ${\bf 0.86}$  & ${\bf 0.87}$  & ${\bf 0.81}$  & ${\bf 0.88}$  && ${\bf 0.65}$ & ${\bf 0.69}$  & ${\bf 0.66}$  & ${\bf 0.67}$  && ${\bf 0.43}$ & ${\bf 0.41}$  & ${\bf 0.42}$  && ${\bf 0.42}$ & ${\bf 0.09}$ & ${\bf 0.08}$ & ${\bf 0.10}$ & ${\bf 0.08}$ \\
\midrule
DnC         & $0.19$        & $0.10$        & $0.33$        & $0.24$        && $0.28$       & $0.10$        & $0.12$        & $0.10$        && $0.10$       & $0.10$        & $0.10$        && $0.11$       & $0.01$       & $0.01$    & $0.01$      & $0.01$ \\
+ FLANDERS  & ${\bf 0.89}$  & ${\bf 0.84}$  & ${\bf 0.88}$  & ${\bf 0.85}$  && ${\bf 0.66}$ & ${\bf 0.65}$  & ${\bf 0.68}$  & ${\bf 0.66}$  && ${\bf 0.42}$ & ${\bf 0.42}$  & ${\bf 0.42}$  && ${\bf 0.42}$ & ${\bf 0.09}$ & ${\bf 0.09}$ & ${\bf 0.09}$ & ${\bf 0.10}$ \\
\bottomrule
\end{tabular}
}
\end{table*}

\begin{table*}[htb!]
\centering
\caption{Accuracy of the global model using Multi-Krum + FLANDERS over various numbers of malicious clients.}
\label{tab:robustness-1}
\scalebox{1}{
\begin{tabular}{cccccccccccccccccccc}
\toprule
& \multicolumn{4}{c}{\textit{MNIST}} & \phantom{a} & \multicolumn{4}{c}{\textit{Fashion-MNIST}} & \phantom{a} & \multicolumn{4}{c}{\textit{CIFAR-10}} & \phantom{a} & \multicolumn{4}{c}{\textit{CIFAR-100}} \\
\cmidrule(lr){2-5} \cmidrule(lr){7-10} \cmidrule(lr){12-15} \cmidrule(lr){17-20}
        & 0      & 20     & 60     & 80     && 0      & 20     & 60     & 80     && 0      & 20     & 60     & 80     && 0      & 20     & 60     & 80     \\ 
\midrule
GAUSS   & $0.80$ & $0.87$ & $0.89$ & $0.87$ && $0.64$ & $0.65$ & $0.71$ & $0.69$ && $0.44$ & $0.42$ & $0.41$ & $0.38$ && $0.08$ & $0.10$ & $0.12$ & $0.11$ \\
LIE     & $0.80$ & $0.85$ & $0.84$ & $0.90$ && $0.64$ & $0.64$ & $0.70$ & $0.68$ && $0.44$ & $0.44$ & $0.40$ & $0.38$ && $0.08$ & $0.09$ & $0.12$ & $0.10$ \\
OPT     & $0.80$ & $0.85$ & $0.84$ & $0.88$ && $0.64$ & $0.69$ & $0.68$ & $0.72$ && $0.44$ & $0.35$ & $0.40$ & $0.39$ && $0.08$ & $0.10$ & $0.11$ & $0.10$ \\
AGR-MM  & $0.80$ & $0.77$ & $0.88$ & $0.89$ && $0.64$ & $0.66$ & $0.67$ & $0.68$ && $0.44$ & $0.42$ & $0.40$ & $0.40$ && $0.08$ & $0.09$ & $0.10$ & $0.11$ \\
\bottomrule
\end{tabular}
}
\end{table*}

\end{document}